\documentclass{altered_singlecol-new}

\usepackage{blindtext, graphicx}

\usepackage{amsmath,amssymb}
\usepackage{cite}
\usepackage{algorithm}
\usepackage{algorithmic}
\usepackage{caption}
\usepackage{ltablex}
\usepackage{booktabs}
\usepackage[bookmarks=false,draft]{hyperref}
\usepackage{subcaption}
\usepackage{url}
\usepackage{color}
\usepackage{graphicx}
\usepackage{xcolor}
\let\labelindent\relax
\usepackage{enumitem}

\usepackage{array}
\usepackage{multicol}

\usepackage{caption}
\usepackage{booktabs}
\usepackage{float}
\usepackage{etoolbox}

\usepackage[bottom]{footmisc}

\usepackage{booktabs}

\usepackage{natbib}

\usepackage{array}
\newcolumntype{L}[1]{>{\raggedright\let\newline\\\arraybackslash\hspace{0pt}}m{#1}}
\newcolumntype{C}[1]{>{\centering\let\newline\\\arraybackslash\hspace{0pt}}m{#1}}
\newcolumntype{R}[1]{>{\raggedleft\let\newline\\\arraybackslash\hspace{0pt}}m{#1}}
\newcommand{\bx}{\ensuremath{\mathbf{x}}}

\newcommand{\bz}{\ensuremath{\mathbf{z}}}

\newcommand{\by}{\ensuremath{\mathbf{y}}}
\newcommand{\ba}{\ensuremath{\mathbf{a}}}
\newcommand{\tildt}{\ensuremath{\tilde{t}}}

\DeclareMathOperator*{\argmin}{arg\,min}

\definecolor{dgreen}{RGB}{63, 175, 115}

\theoremstyle{TH}{

}

\theoremstyle{THrm}{
\newtheorem{definition}{Definition}[section]

}

\theoremstyle{THhit}{

}

\makeatletter
\def\theequation{\arabic{equation}}

\makeatother

\begin{document}

\title{Deriving Enhanced Geographical Representations via Similarity-based Spectral Analysis: Predicting Colorectal Cancer Survival Curves in Iowa}

\setcounter{page}{1}

\LRH{MT. Lash et~al.}

\RRH{Deriving Enhanced Geographical Representations}

\VOL{x}

\ISSUE{x}

\PUBYEAR{2018}

\BottomCatch

\CLline

\subtitle{}

\authorA{Michael T.~Lash}
\affA{Department of Computer Science,\\ University of Iowa,\\ Iowa City, IA, USA \\
\qquad E-mail: michael-lash@uiowa.edu}
\authorB{Min Zhang}
\affB{Interdisciplinary Graduate Program in Informatics,\\ University of Iowa,\\ Iowa City, IA, USA \\
\qquad E-mail: min-zhang@uiowa.edu}

\authorC{Xun Zhou}
\affC{Management Sciences Department,\\ University of Iowa,\\ Iowa City, IA, USA \\
\qquad E-mail: xun-zhou@uiowa.edu}

\authorD{W.~Nick Street}
\affD{Management Sciences Department,\\ University of Iowa,\\ Iowa City, IA, USA \\
\qquad E-mail: nick-street@uiowa.edu}

\authorE{Charles F.~Lynch}
\affE{Department of Epidemiology,\\ University of Iowa,\\ Iowa City, IA, USA \\
\qquad E-mail: charles-lynch@uiowa.edu}

\begin{abstract}
Neural networks are capable of learning rich, nonlinear feature representations shown to be beneficial in many predictive tasks. In this work, we use such models to explore different geographical feature representations in the context of predicting colorectal cancer survival curves for patients in the state of Iowa, spanning the years 1989 to 2013. Specifically, we compare model performance using \textit{area between the curves} (ABC) to assess (a) whether survival curves can be reasonably predicted for colorectal cancer patients in the state of Iowa, (b) whether geographical features improve predictive performance, (c) whether a simple binary representation, or a richer, spectral analysis-elicited representation perform better, and (d) whether spectral analysis-based representations can be improved upon by leveraging geographically-descriptive features. In exploring (d), we devise a similarity-based spectral analysis procedure, which allows for the combination of geographically relational and geographically descriptive features. Our findings suggest that survival curves can be reasonably estimated on average, with predictive performance deviating at the five-year survival mark among all models. We also find that geographical features improve predictive performance, and that better performance is obtained using richer, spectral analysis-elicited features. Furthermore, we find that similarity-based spectral analysis-elicited representations improve upon the original spectral analysis results by approximately 40\%.
\end{abstract}

\KEYWORD{Geographical representations; Spectral analysis; Deep learning; Spectral clustering; Neural networks; Colorectal cancer; Survival curve}

\REF{to this paper should be made as follows: Lash, M.T., Zhang, M.~, Street, W.N., Zhou, X., and Lynch, C.F. (2018) `Deriving Enhanced Geographical Representations via Similarity-based Spectral Analysis: Predicting Colorectal Cancer Survival Curves in Iowa', {\it arXiv preprint}, 2018.}

\begin{bio}
Michael Lash is currently a PhD candidate in the Department of Computer Science at the University of Iowa, advised by W.~Nick Street and Alberto M.~Segre. His research interests are broadly in the areas of machine learning and data mining methodology, with specific interests lying in causal learning, learning from graphs, geographical and spatial data mining, among others. He has published in top data mining conferences, such as SDM, top application venues, such as ICHI and BIBM, and top business journals, such as JMIS. He has also been the recipient of numerous awards, including the University of Iowa Graduate College Post-Comprehensive Research Fellowship, an NSF GRFP Honorable Mention, and a variety of student travel awards.

\noindent Min Zhang is currently a graduate student in the Interdisciplinary Graduate Program in Informatics at the University of Iowa. He received the bachelor degree in Business Analytics and Information Systems from the University of Iowa, in 2017. His research interests include business data analytics, big data management, and Geographic Information Systems (GIS).

\noindent Xun Zhou is currently an Assistant Professor in the Department of Management Sciences at the University of Iowa. He received a PhD degree in Computer Science from the University of Minnesota, Twin Cities in 2014. His research interests include big data management and analytics, spatial and spatio-temporal data mining, and Geographic Information Systems (GIS). His works have been published in top conferences and journals such as ACM SIGKDD, IEEE ICDM, ACM SIGSPATIAL, and IEEE TKDE. Xun has received three best paper awards. He was also a co-editor-in-chief of the Springer Encyclopedia of GIS, 2nd Edition.

\noindent Nick Street is the Henry B. Tippie Research Professor and Departmental
Executive Officer in the Management Sciences Department at the
University of Iowa, with joint appointments in Computer Science,
Nursing, and Informatics.  He is also the director of the
interdisciplinary graduate program in Health Informatics.
His research interests are in algorithmic approaches to machine
learning and data mining, particularly the use of mathematical
optimization in inductive learning techniques.  His recent work has
focused on counterfactual reasoning, ensemble construction methods, 
and personalized health care decision making.  He has
published over 110 journal, conference and workshop papers, and has
received an NSF CAREER award and an NIH INRSA postdoctoral fellowship.

\noindent Charles Lynch is a Professor with a joint appointment in the Department of Epidemiology in the College of Public Health and in the Department of Pathology in the College of Medicine at The University of Iowa.  He has been a faculty member at The University of Iowa since he completed his pathology training in 1986.  Since 1990, he has been Principal Investigator of the State Health Registry of Iowa, Iowa's statewide cancer surveillance program.  His primary research interests include cancer surveillance, cancer epidemiology, and environmental epidemiology.

\end{bio}

\maketitle

\section{Introduction}

As machine learning has become more prevalent, powerful new technologies such as deep learning, which are capable of learning rich, non-linear representation, have also risen to the forefront of the field. The domains of public health and medicine have particularly benefited from these innovations; in this work we examine and propose deep learning methodologies applied to these areas. The focus of this work, therefore, is to explore how different geographical representations, learned through deep learning technologies, can improve survival curve predictions for colorectal cancer patients in the state of Iowa.

\begin{figure}[h]
    \centering
    \includegraphics[scale=.40]{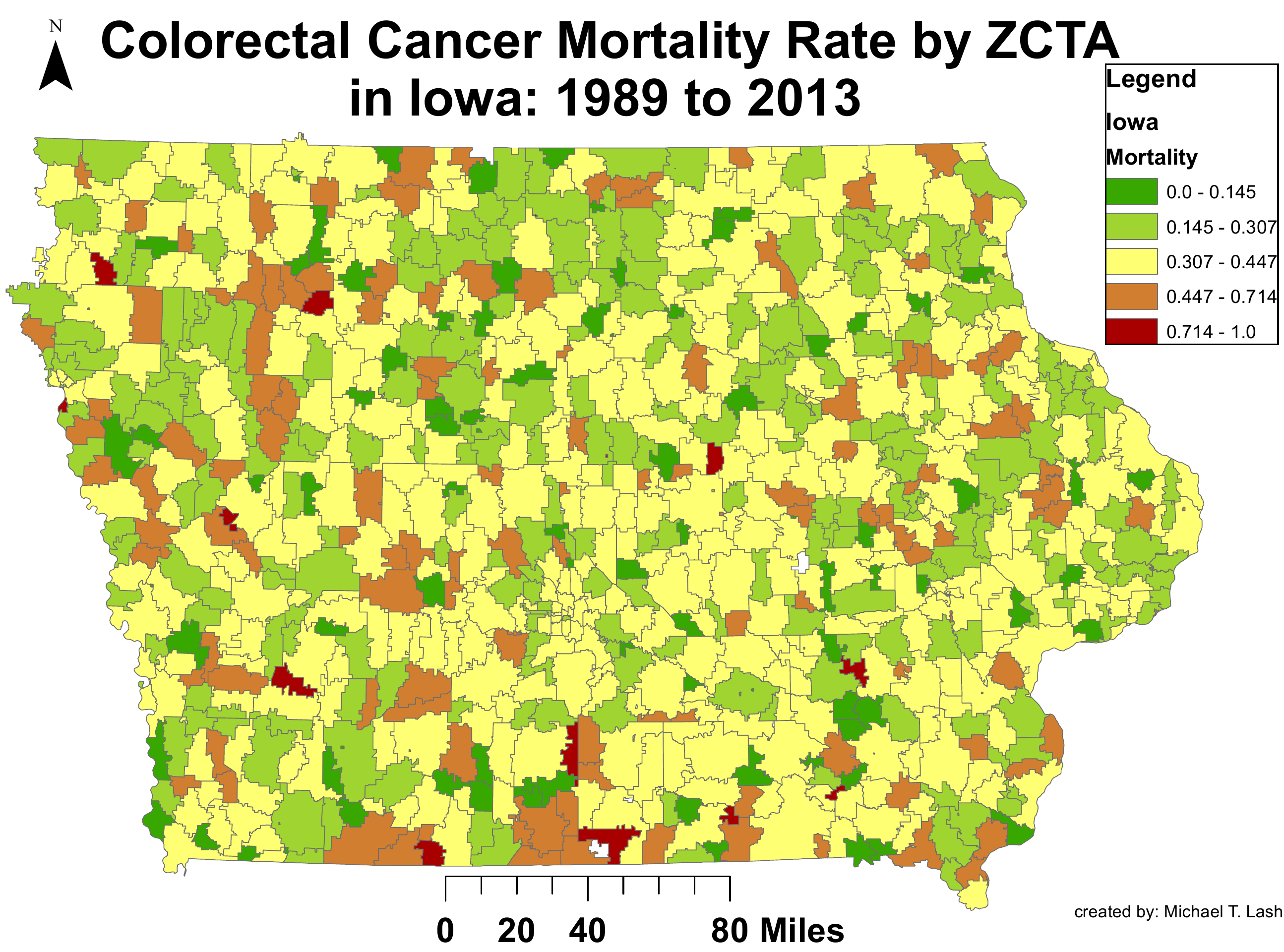}
    \caption{Colorectal cancer mortality rate by ZCTA in the state of Iowa for the years 1989 to 2013.}
    \label{fig:mortrate}
\end{figure}

Figure \ref{fig:mortrate} demonstrates the urgency of the problem we are addressing, showing colorectal cancer (CRC) mortality rates for patients in Iowa spanning the years 1989 to 2013; these are expressed in terms of a zipcode tabulation area (ZCTA) level of geography.
In Figure \ref{fig:mortrate} we first observe that numerous ZCTAs have CRC mortality rates that are at or above the 30\%, indicating the particularly nefarious nature of this disease, and highlighting the need for accurate survival outlook predictions at the time of diagnosis to better inform treatment decisions \cite{Zhang2015DM}. Furthermore, Figure \ref{fig:mortrate} demonstrates the geographical diversity in which CRC mortality rates are manifested: locale seems to be related to survival outlook.

The relationship between geography and survival outlook isn't unforeseen, unfortunately. Physical locale manifests pertinent health-based factors, such as access to health care, environmental factors, such as ground contaminants, among others, all of which may affect disease manifestation and survival outlook \cite{wan2013spatial}.

Provided the spatially heterogeneous manifestation of colorectal cancer mortality, a major challenge is to build spatially responsive models that can aid in accurate prediction of individual-specific colorectal cancer survival curves. For instance, rural areas may have a different variety of factors affecting colorectal cancer disease manifestation and survival than sprawling metropolitan cities. Therefore, we define and examine three geographical deep learning representation methods in this work: a \textit{simple binary representation} (SBR), \textit{rich representation -- spectral analysis} (RR-SA), and rich representation --similarity-based spectral analysis (RR-SSA); we additionally craft two sub-representation methods that are utilized in RR-SSA.

The contributions of this work, which expand upon the results obtained in \cite{lash2017learning}, are enumerated as follows:
\begin{enumerate}
    \item We investigate a rich representation of spatial features through spectral analysis (RR-SA) of the underlying geographical relationship graph of the ZCTAs to address the spatial heterogeneity challenge. 
    \item Modifying our RR-SA representation procedure, we explore the use of geographically descriptive features, paired with the underlying adjacency graph, to further address the spatially heterogeneous nature of the problem.
    \item We determine whether the simple binary representation (SBR) or richer, spectral analysis representation (RR-SA), or similarity-based spectral analysis representation (RR-SSA) leads to more accurate survival curve predictions.
    \item We determine whether RR-SA or RR-SSA representations lead to more accurate survival curve predictions and determine which sub-representation procedure -- binary (bin) or full -- produces more accurate survival curve predictions.
\end{enumerate}

This works continues with a disclosure of the problem setting, followed by relation of our three methods of geographic representation and two sub-representation methods; we also present a graphic depicting the deep architecture of each method (Section 2). In Section 3, we describe our colorectal cancer patient dataset, containing ~46000 individuals residing in Iowa at the time of their diagnosis; the dataset spans the years 1989 to 2013. Furthermore, we relate our geographical feature dataset, along with our experiments. In Section 4 we disclose works related to ours prior to concluding the paper in Section 5.

\section{Learning Geographical Representations for Survival Curve Prediction}

Prior to disclosing our methodology, we relate some preliminary notation, subsequently discussing and mathematically formulating the problem setting. Following this disclosure we reformulate the problem as one of Kaplan-Meier survival curve prediction before introducing and elaborating on our three methods of geographical representation learning and two sub-representations.

\subsection{Preliminaries}

Define $\{(\bx^{(i)}, e^{(i)}, t^{(i)})\}_{i=1}^{n}$ to be a dataset of $n$ instances, where feature vector $\bx^{(i)} \in \mathbb{R}^{m}$, event label $e^{(i)} \in \{0,1\}$, and time of event occurrence $t^{(i)} \in \{0,1,\dots,T\}$; where $t^{(i)}$ represents a discrete time at which an event $e^{(i)}$ has occurred (i.e.,~$e^{(i)}=1$) \textit{or} the last discrete time instance $i$ is observed, while an event has not occurred (i.e.,~$e^{(i)}=0$). In this latter case ($e^{(i)}=0$), when $t^{(i)} = T$ we know the event never occurs to the instance during the study period (spanning $T$ discrete time periods). If, on the other hand, $t^{(i)} < T$ then we only know that the instance did not experience the event up to $t^{(i)}$, but don't know what happened during the $T-t^{(i)}$ remaining time. Representation of event-time data described as such are called \textit{censored data} and, even more specifically, \textit{right-censored} data. A censoring of instance $i$ occurs when $e^{(i)}=0$ and $t^{(i)} < T$. We elaborate on the handling of these censored data in a proceeding subsection.

To be more concrete, $t \in \{1,\dots,T\}$ may represent (as is the case in our experiments) six-month patient follow-up periods, with $t=0$ designating the entrance of a patient to the study. Study entrance occurs when a diagnosis of colorectal cancer is rendered. When an instance (i.e.,~patient) $i$ dies from colorectal cancer -- $e^{(i)}=1$ -- then  $t^{(i)}$ designates a time in which this event occurred. Alternately, a patient may pass away from complications not related to their colorectal cancer disease, or may move elsewhere, switch doctors, or for some other reason become untrackable prior to the conclusion of the study period, then $t^{(i)} < T$ and $e^{(i)}=0$, indicating a censoring.

Patient instance vectors $\bx^{(i)}$ represent quantified measurements of pertinent patient-based features. Later in this work, we will make reference to certain feature groups of which these instance vectors are composed. Therefore, we define notation that will conveniently relate to these groups. To such an end, let $\bz$ denote the full set of index values that reference the geographical features of $\bx^{(i)}$; further, denote $\ba$ to be the full set of index values of $\bx^{(i)}$ such that $\ba = \{1,\dots,m\}$. We will use these index sets to make direct reference to the feature grouping components of $\bx^{(i)}$; for instance, $\bx^{(i)}_{\bz}$ is the subvector of instance $i$ housing the geographical feature values. Furthermore, using set difference notation, $\bx^{(i)}_{\ba \setminus \bz}$ refers to the subvector of instance $i$ containing feature values that are non-geographical.
\begin{table}[h]
\centering
\begin{tabular}{ll}
\toprule
\textbf{Notation} & \textbf{Description} \\ \midrule
$\bx^{(i)} \in \mathbb{R}^m$ & Feature vector of instance $i$. \\
$e^{(i)} \in \{0,1\}$ & Event label of instance $i$. \\
$t^{(i)} \in \{1,\dots,T\}$ & Discrete time of $e^{(i)}$. \\
$\by^{(i)} \in [0,1]^{T}$ & Outcome vector of instance $i$. \\ 
$\hat{\by}^{(i)} \in [0,1]^{T}$ & Predicted outcome vector of instance $i$.\\\midrule
$\bz$ & Set of geographical feature index values.\\
$\ba$ & Set of all feature index values.\\
$\mathcal{M}$ & A map.\\
$\Gamma(\cdot)$ & Function that determines discrete\\
& geographic entity membership.\\\midrule
$P(\cdot)$ & Calculation of a probability.\\
$\mathtt{g}:\mathbb{R}^m \rightarrow [0,1]^{T}$ &  Neural network.\\
$\mathcal{L}(\cdot)$ & An arbitrary loss function.\\
$\mathtt{Smooth}$ & Output smoothing function.\\\midrule
$\pmb{\mathbb{Z}}$ & Adjacency matrix constructed from $\mathcal{M}$. \\
$\hat{\pmb{\mathbb{Z}}}$ & SSA-elicited affinity matrix.\\
$\pmb{\mathbb{A}}$ & Design matrix for geographical\\
& entities (descriptive geo feats).\\
$\mathtt{Common}$& Function that determines whether two geographic entities in\\
&$\mathcal{M}$ are adjacent.\\
$\pmb{Q}_{spec}$& Top $k$ eigenvectors from $\pmb{Q}$, selected based on largest\\
& eigenvalues in $\pmb{\lambda}$.\\
$\mathbf{q}_{label}$& The result of applying kMeans clustering to $\pmb{Q}_{spec}$.\\
$\mathtt{Enrich}$& Function that assigns values in $\pmb{Q}_{spec}$ to an instance.\\
$\pmb{\Theta}$ & SSA procedure to produce $\hat{\pmb{\mathbb{Z}}}$.\\\bottomrule
\end{tabular}
\caption{Notation used throughout this work.\label{tab:notation}}
\end{table}

For convenience, we provide the notation related in this and subsequent sections in Table \ref{tab:notation}.

\subsection{Kaplan-Meier Re-representation}

To begin elaborating on the censored nature of our data, as we mentioned in the previous section, instance $i$ has an event label $e^{(i)}$ and a discrete time of event occurrence $t^{(i)}$: provided this, the goal is to transform this two-valued representation to that of a \textit{Kaplan-Meier survival curve} (KMSC) representation \cite{kaplan1958nonparametric}. A KMSC, simply put, each temporal unit $1,\dots, T$ with a probability of the disease event $e^{(i)}$ \textit{not} occurring at that particular temporal unit, dependent upon the probability of ``not'' event occurrence of the preceding temporal unit, for each instance $i$. 

More formally, the KMSC re-representation is in the form of a vector, denoted $\by^{(i)} \in [0,1]^{T}$, where the index values $\tildt \in \{1,\dots,T\}$ express the temporal units and the indexed values $\by^{(i)}_{\tildt}$ denote the respective probabilities.

Our KMSC re-representation scheme is originally outlined in Chi et al.~\cite{chi2007application}. To instantiate the vector $\by^{(i)}$, the following is conducted:
\begin{align}
    \label{eq:rerep}
    y^{(i)}_{\tildt} = \left\{
    \begin{array}{ll}
    1 & \text{if }\tildt < t^{(i)}\\
    0 &\text{if }\tildt \geq t^{(i)} \text{ \& } e^{(i)} = 1\\
    1 - P(e_{\tildt}^{(i)}=1 | e_{\tilde{t}-1}^{(i)}=0) & \text{if }\tilde{t} \geq t^{(i)} \text{ \& } e^{(i)} = 0
    \end{array}
    \right.
\end{align}
where $P(e_{\tildt}^{(i)}=1 | e_{\tildt-1}^{(i)}=0)$ denotes the conditional probability of event $e^{(i)}$ occurring at $\tildt$ provided that $e^{(i)}$ has not occurred at $\tildt - 1$. As such, for patients whose CRC outcomes are known, $\by^{(i)}$ exhibits values that are strictly 0 and 1. On the other hand, a censored patient's vector exhibits estimation of survival probability beginning at the index position $\tilde{t} = t^{(i)}$; the ensuing values are conditional probability estimates.

\subsection{Predicting Individual KMSC}

Our goal in this work is to induce an optimal hypothesis $\mathtt{g}^* \in \mathcal{G}$ of some [presently] arbitrarily defined hypothesis class $\mathcal{G}$, that is most apt at predicting instance-specific KMSCs. We formalize this problem as:
\begin{align}
    \mathtt{g}^*=\argmin\limits_{\mathtt{g} \in \mathcal{G}} \left\{ \mathcal{L}\left(\by^{(i)},\mathtt{g}(\bx^{(i)})\right): i=1,\dots,n \right\}
\end{align}
where $\mathcal{L}(\cdot)$ expresses some loss function that measures the divergence between the predicted $\by^{(i)}$ (henceforth expressed $\hat{\by}^{(i)}$) and the known $\by^{(i)}$. 

The hypothesis class $\mathcal{G}$ explored in this work is defined as both deep and shallow neural network architectures, the specifics of which are disclosed later in this section; we discuss the specific parameterizations employed across our experiments in the experiments section (Section 3). Deep neural network architectures are characterized by multiple hidden layers, and shallow architectures by a single hidden layer.

\subsubsection{Output Smoothing}

Construction of a neural network model is accomplished in layer-wise fashion, where a particular layer is composed of nodes. The first layer in a neural network is designated as the input layer, which is proceeded by any number of so-called hidden layers, the last of which is connected to the output layer. The output layer is somewhat unique to our problem setting of predicting KMSCs. First, the predicted probability elicited from each of the $\tilde{t} = 1,\dots,T$ output nodes are \textit{ordered}. In other words, the output of $node_{\tilde{t}}^{out}$ is \textit{ordered} before $node_{\tilde{t}+1}^{out}$ because $\tilde{t}$ is temporally occurs before $\tilde{t}+1$. Second, the ordered output probabilities of these nodes should be strictly decreasing: i.e.,~$output_{\tilde{t}}^{(i)} \geq output_{\tilde{t}+1}^{(i)}$. The reasoning behind this ``strictly decreasing'' expectation is intuitive: the probability of survival, of a disease or otherwise, even after disease recovery, is never expected to go up. The loss function $\mathcal{L}(\cdot)$ employed to induce mulitple-output networks, such as those in our problem setting, elicit a single loss value representing the loss across all nodes, meaning the desired strictly decreasing output cannot be guaranteed. In light of this, we develop a smoothing operation, denoted $\mathtt{Smooth}(\mathbf{output}^{(i)})$, formally expressed by
\begin{align}
\label{eq:smooth}
\hat{y}^{(i)}_{\tilde{t}+1} = \min\{output_{\tilde{t}}^{(i)},output_{\tilde{t}+1}^{(i)}\} \text{ for } \tilde{t}=1,\dots,T
\end{align}
guaranteeing that the post-processed (i.e.,~smoothed) model output is strictly decreasing.

\subsection{Geographic Feature Representation}

Although our primary concern is to elicit a $\mathtt{g}^{*}$ that produces the most accurate predictions, the novelty of the work is to:
\begin{enumerate}
\item Demonstrate that geographic-based feature representations enhance the quality of predictions.
\item Explore whether simple binary representations or a richer representations (defined shortly) produce more accurate predictions.
\item Quantify the extent to which these representations improve predictive quality.
\end{enumerate}
The details of our experiments and data are elaborated on in the next section where we explore three different geographical representations: a simple binary representation (SBR), a rich representation based on spectral analysis (RR-SA), and a rich representation employing similarity-based spectral analysis (RR-SSA).

\subsubsection{Simple Binary Representation}

Our simple binary representation (SBR) is minimalist in nature, the procedure consisting only of (a) determination of the discrete geographic entity membership of instance $i$ and (b) such membership being binarily re-represented (referred to as \textit{one hot encoding}), thus eliciting a sparse vector-based encoding with a $1$ in the indexical location referring to the geographic entity of which $i$ is a member, and $0$s in the remaining positions. 

To devise a formulation that is aptly generalizable we assume that the geographic features of instance $i$, expressed as $\bx^{(i)}_{\bz}$, are defined such that encoded values are capable of eliciting the discrete geographic unit of which $i$ is a member (e.g.,~coordinates). For instance, we employ ZCTAs (zipcode tabulation area) as the discrete geographic unit in our experiments.

A formal procedure for eliciting discrete geographic unit membership can be expressed as
\begin{align}
\label{eq:geomem}
    x^{(i)}_{b} = \Gamma(\bx^{(i)}_{\bz},\mathcal{M})
\end{align}
where the function $\Gamma(\cdot)$ performs a transformation on $\bx^{(i)}_{\bz}$, the geographic feature values of the instance, to some identification (ID) value, which we denote as $x^{(i)}_{b}$. This $x^{(i)}_{b}$ value denotes the unique, discrete geographic entity, belonging to map $\mathcal{M}$ (which we define momentarily), of which instance $i$ is a member. The values represented by $\bx^{(i)}_{\bz}$, along with the information expressed in map $\mathcal{M}$, dictate the procedure used by $\Gamma(\cdot)$ to perform the transformation.

The specific $\bz$ geographics features employed in this work are (latitude,longitude) coordinate pairs and, as such, we specify a definition (referred to as Definition \ref{def:map}) of map $\mathcal{M}$ using geography defined in terms of these coordinate pairs.
\begin{definition}
\label{def:map}
Define $\mathcal{M}$ to be a \textbf{map}, given by
\begin{align}
    \mathcal{M} = \left \{(key_l,value_l)\right\}_{l=1}^{p}
\end{align}
where $key_l$ is the unique postal code of geographic unit $l$ and $value_l$ is an ordered set of (lat,lon) coordinate pairs denoting the bounding geographic region of $l$.

We characterize map $\mathcal{M}$ as a continuous geographic region by
\begin{align}
    \left\{ \forall \mathtt{l} \exists \mathtt{l}^{\prime}: value_{\mathtt{l}}^{q} = value_{\mathtt{l}^{\prime}}^{j} \text{ for } \mathtt{l},\mathtt{l}^{\prime} \in \{1,\dots,p\} \text{ \& } \mathtt{l} \neq \mathtt{l}^{\prime}
    \right\}
\end{align}
where $value_{\mathtt{l}}^{q} = value_{\mathtt{l}^{\prime}}^{j} \triangleq (lat_{\mathtt{l}}^q = lat_{\mathtt{l}^{\prime}}^j) \cap (lon_{\mathtt{l}}^q = lon_{\mathtt{l}^{\prime}}^j$).
\end{definition}


Provided our definition of map $\mathcal{M}$, expressed in Definition \ref{def:map}, we define  $\Gamma(\cdot)$ to be a function that takes (lat,lon) coordinate pairs $\bx^{(i)}_{\bz}$ and determines whether the point is on the interior of each ZCTA. The zipcode ID, corresponding to the ZCTA having the point $\bx^{(i)}_{\bz}$ on the interior, is subsequently initialized as the value of $x^{(i)}_{b}$ (i.e.,~$x^{(i)}_{b}=\text{ZCTA}_{text{ID}}$). Following this outlined $\Gamma(\cdot)$ procedure, and additional binarization procedure, often referred to as one-hot encoding, which we denote $\mathtt{Bin}$, is employed to produce a spare vector representation consisting of a single $1$ in the position referencing the ZCTA of which instance $i$ belongs, and $0$s in other positions.

\begin{figure}[h]
    \centering
    \includegraphics[scale=.60]{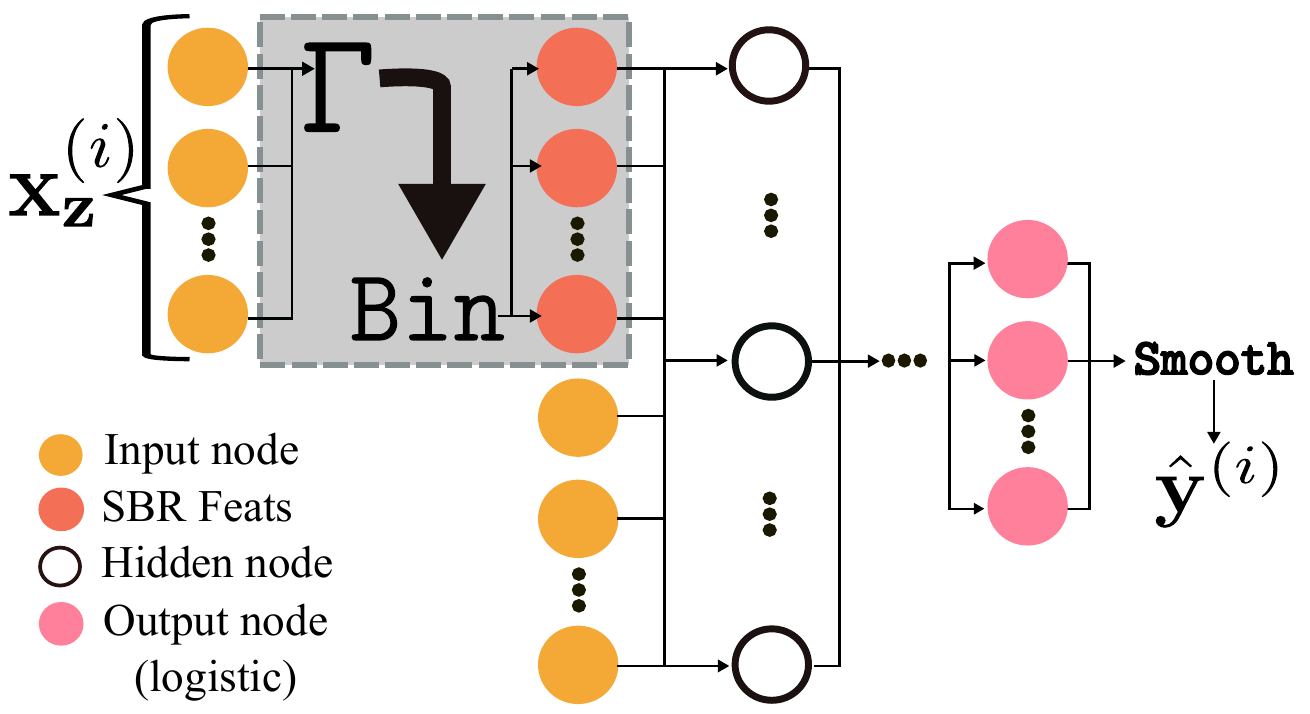}
    \caption{SBR neural network architecture.}
    \label{fig:sbr-arch}
\end{figure}

Figure \ref{fig:sbr-arch} illustrates the network architecture using the SBR methodology.

We expect that non-geographic representations will perform worse than representations employing the SBR representation

While models elicited from employing the SBR representation may enjoy some predictive performance improvement over hypotheses induced on strictly non-geogrpahic features, representations that consist of richer geographic encodings, capable of modeling the continuous nature of the geographic region of study promise to produce even better results.

\subsubsection{Spectral Analysis Representation}
To elicit richer geographical representations, we devise a spectral analysis approach, based on on a well-known procedure referred to as spectral clustering. The method begins by first computing an adjacency matrix among the discrete geographic entities represented in $\mathcal{M}$. Subsequently, spectral analysis solves for the eigenvectors and eigenvalues of the adjacency representation, selecting the top $k$ eigenvectors, based on the largest $k$ eigenvalues. The elicited representation is $p \times k$ matrix, where the $p$ rows refer to the $p$ discrete geographic entities (i.e., a single row refers to one of the $p$ entities). The $k$ values that compose each row are used as geographic predictive input features.

To more formally relate this spectral analysis procedure, define $\pmb{\mathbb{Z}} = \mathtt{Adj}\left(\mathcal{M}\right)$ to be the affinity (i.e.,~adjacency, similarity) matrix, in which the $l,v$-th entry relates the geographic adjacency relationship among the $l$-th and $v$-th entities. We express this by
\begin{align}
    \label{eq:adjmatform}
    [\pmb{\mathbb{Z}}]_{l,v} = \left\{
    \begin{array}{ll}
    1 & \text{if }\ \mathtt{Common}(values_l,values_v) = True \\
    & \text{\& } l \neq v\\
    0 & \text{otherwise}
    \end{array}
    \right.
\end{align}
where the function $\mathtt{Common}(\cdot)$ determines if $values_l$ and $values_v$ have a common element. Provided $\mathcal{M}$, related by Definition \ref{def:map}, $\mathtt{Common}(\cdot)$ computes whether or not $values_l$ and $values_v$ share at least one coordinate pair.

Subsequently, spectral clustering is executed by performing $k$Means clustering, $\mathbf{q}_{label}= kMeans \left(\pmb{Q}_{spec}\right)$, where the function $kMeans(\cdot)$ assigns one of the $k$ cluster labels to each of the $p$ entries of $\pmb{Q}_{spec}$; where
\begin{align}
\label{eq:qspec}
\pmb{Q}_{spec} = \mathtt{Top}_k\left(\pmb{Q},\pmb{\lambda}\right).
\end{align}
The function $\mathtt{Top}_k(\cdot)$ searches and finds the largest values in $\pmb{\lambda}$, selects the appropriate columns in the matrix $\pmb{Q}$, and creates the matrix $\pmb{Q}_{spec} \in \mathbb{R}^{k\times p}$. The matrix $\pmb{Q}$, composed of eigenvectors, and corresponding vector $\pmb{\lambda}$, composed of eigenvalues, are obtained by solving the system of equations related by
\begin{align}
\label{eq:sceiq}
\pmb{\mathbb{Z}}\pmb{Q} = \pmb{\lambda} \pmb{Q} .
\end{align}
Here, the columns of $\pmb{Q}_{spec}$ are used as the $k$ geographical features when inducing a hypothesis $\mathtt{g}$ -- we refer to this use of the $\pmb{Q}_{spec}$ matrix as spectral analysis. The labels, $\mathbf{q}_{label}$, obtained from application of the clustering procedure, referred to as spectral clustering, are used to visualize the elicited representations obtained from our experiments, related in the next section. Spectral analysis avoids making use of binarized label assignments of the spectral clustering procedure, instead using a subprocedure, termed spectral analysis, which preserves cluster composition and is a richer (i.e.,~non-sparse) representation.

In Algorithm \ref{algo:sc}, we relate the spectral clustering process, differentiating spectral analysis from spectral clustering via \textcolor{red}{red} highlighting; omission of this line produces the spectral analysis procedure.
\begin{algorithm}
    \caption{Spectral Clustering\label{algo:sc}}
    \begin{algorithmic}[1]
    \STATE Obtain adjacency matrix $\pmb{\mathbb{Z}}$ using \eqref{eq:adjmatform}.
    \STATE Solve \eqref{eq:sceiq} for $\pmb{Q}$ and  $\pmb{\lambda}$.
    \STATE Obtain $\pmb{Q}_{spec}$ as outlined in \eqref{eq:qspec}.
    \STATE \textcolor{red}{Apply kMeans clustering to }$\color{red}\pmb{Q}_{spec}$ \textcolor{red}{to obtain }$\color{red}\mathbf{q}_{label}$\textcolor{red}{.}
    \end{algorithmic}
\end{algorithm}
Simply put, spectral analysis is a sub-procedure of the spectral clustering process, yielded by omitting the clustering step.

\begin{figure}[h]
    \centering
    \includegraphics[scale=.60]{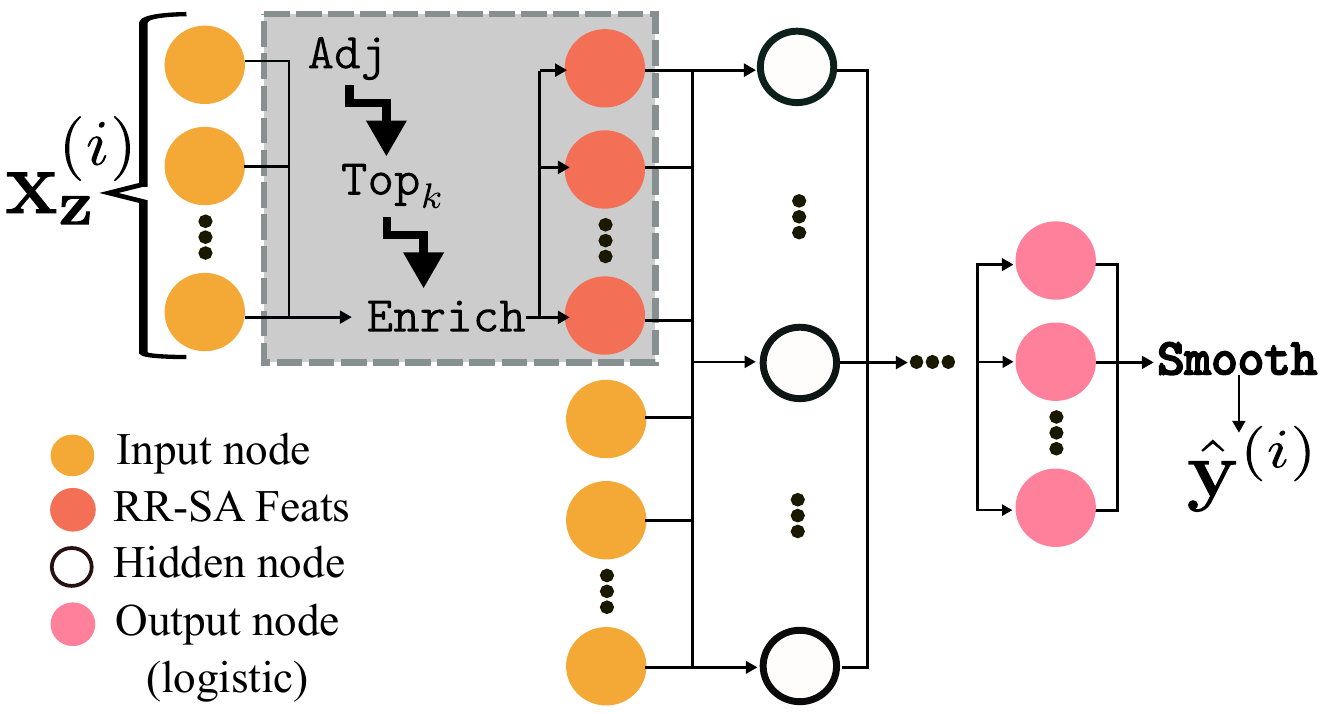}
    \caption{RR-SA neural network architecture.}
    \label{fig:rr-sc-arch}
\end{figure}

Finally, for a test instance $\bx$, a process $\mathtt{Enrich}(\bx_{\bz},\mathcal{M},\pmb{Q}_{spec})$ is executed to obtain the appropriate $k$-valued column entry of $\pmb{Q}_{spec}$ that is associated with the particular geographic entity that the test instance belongs to. Algorithm \ref{algo:Enrich} outlines this procedure.
\begin{algorithm}
	\caption{Enrich Geographic Features $\pmb{\mathtt{Enrich}}$}
	\label{algo:Enrich}
	\begin{algorithmic}[1]
	    \REQUIRE $\bx_{\bz},\mathcal{M},\pmb{Q}_{spec}$
	    \STATE $x_b = \Gamma(\bx_{\bz},\mathcal{M})$ From \eqref{eq:geomem}.
	    \STATE Using $x_b$ find the $l$ such that $x_b = key_l: l \in \{1,\dots,p\}$.
	    \ENSURE Return column vector $[\pmb{Q}_{spec}]_l$
	\end{algorithmic}
\end{algorithm}
The deep learning network architecture outlining the spectral analysis procedure in conjunction with learning a hypothesis, is depicted in Figure \ref{fig:rr-sc-arch}\footnote{As previously mentioned,  $\mathbf{x}_{\mathbf{z}}^{(i)}$ denote (latitude,longitude) coordinate pairs.}.

\subsubsection{Similarity-based Spectral Analysis Representation}

While the above spectral analysis-based approach produces richer geographic representations, the process is capable of leveraging only geographically relational information, as the input affinity matrix must be square (i.e.,~$p \times p$).
It may, however, be beneficial to leverage encodings that are both relational and descriptive in nature, as geographically-descriptive features, such as population demographics and types of land-use, may further enrich the spectral analysis-elicited representations. 

To allow for such input matrices we adjust our spectral analysis process to first calculate the pairwise similarity among geographic entities, thus producing a square affinity matrix on which the spectral analysis procedure can be performed.

More formally, recall the previously discussed adjacency matrix $\pmb{\mathbb{Z}}$ and define $\pmb{\mathbb{A}} \in \mathbb{R}^{p \times h}$ to be a geographic entity design matrix whose rows represent geographic entities and whose columns represent features. Additionally, $\pmb{\mathbb{A}}$ is constructed such that the $l$th row of $\pmb{\mathbb{Z}}$ and the $l$th row of $\pmb{\mathbb{A}}$ refer to the same entity.

Using $\pmb{\mathbb{A}}$ and $\pmb{\mathbb{Z}}$, along with a similarity measure, we devise two sub-representation methods from which a $p \times p$ affinity matrix can be derived.

The first sub-representation uses a single binary feature to indicate spatial adjacency between two geographic entities along with the geographically descriptive features of each entity to yield two vectors $\hat{\mathbf{z}}_l$ and $\hat{\mathbf{z}}_v$. These can be formally expressed as
\begin{align}
    \label{eq:sub1}
    \hat{\mathbf{z}}_l = & [\pmb{\mathbb{Z}}_{l,v},\pmb{\mathbb{A}}_{l}] \\ \nonumber
    \hat{\mathbf{z}}_v = & [\pmb{\mathbb{Z}}_{v,l},\pmb{\mathbb{A}}_{v}]
\end{align}
where $[ \cdot ]$ represents the concatenation of a scalar or vector with another vector (in this case, it is scalar-vector concatenation). Also, note that $\pmb{\mathbb{Z}}_{l,v} = \pmb{\mathbb{Z}}_{v,l}$. We term this sub-representation method \textit{SSA (bin)}.

The second sub-representation uses full geographic entity adjacency vectors instead of a definitive indicator of immediate spatial adjacency. This sub-representation method can be expressed by
\begin{align}
    \label{eq:sub2}
    \hat{\mathbf{z}}_l = & [\pmb{\mathbb{Z}}_{l},\pmb{\mathbb{A}}_{l}] \\ \nonumber
    \hat{\mathbf{z}}_v = & [\pmb{\mathbb{Z}}_{v},\pmb{\mathbb{A}}_{v}]
\end{align}
where, in this case $[ \cdot ]$ indicates two vectors being concatenated.  We term this sub-representation method \textit{SSA (full)}.

Subsequently, using either SSA (bin) or SSA (full), the cosine similarity, denoted as $\phi(\cdot)$, between $l$ and $v$ is computed, thus producing an affinity matrix.

\begin{figure}[h]
    \centering
    \includegraphics[scale=.60]{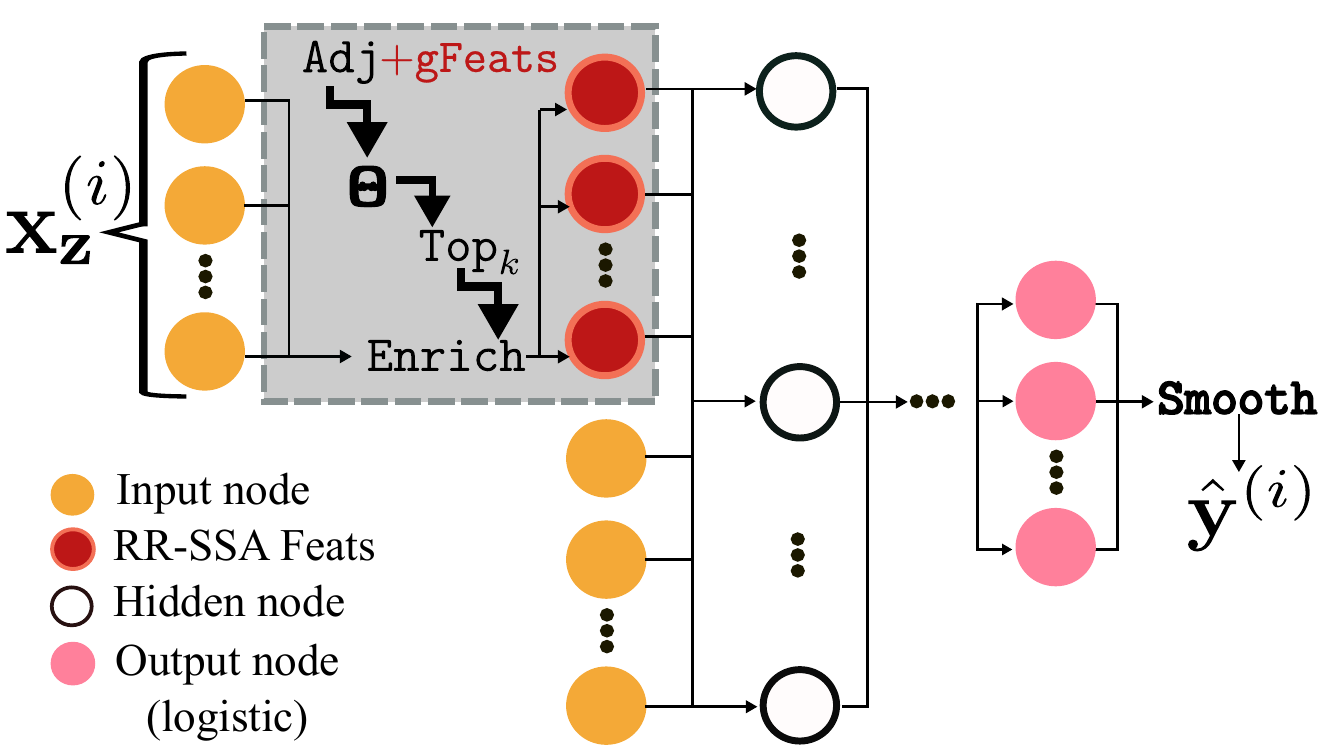}
    \caption{RR-SSA neural network architecture.}
    \label{fig:rr-ssa-arch}
\end{figure}

Algorithm \ref{algo:Theta}, which denotes the procedure as $\pmb{\Theta}$, fully discloses this process, using either SSA (bin) or SSA (full), while Figure \ref{fig:rr-ssa-arch} expresses the process in the context of the neural network architecture.

\begin{algorithm}[h]
	\caption{Sub-rep to affinity $\pmb{\Theta}$}
	\label{algo:Theta}
	\begin{algorithmic}[1]
	    \REQUIRE $\pmb{\mathbb{A}},\pmb{\mathbb{Z}}, \mathtt{REP} \in \{\text{SSA (bin)},\text{SSA (full)}\}$
	    \STATE $\hat{\pmb{\mathbb{Z}}} \leftarrow \mathbf{0}^{p \times p}$
	    \FOR{$l=1,\dots,p-1$}
	    \FOR{$v=l+1,\dots,p-1$}
	    \IF{$\mathtt{REP}=$SSA (bin)}
	        \STATE Define $ \hat{\mathbf{z}}_l$ and $\hat{\mathbf{z}}_v$ according to \eqref{eq:sub1}.
	    \ELSE
	        \STATE Define $\hat{\mathbf{z}}_l$ and $\hat{\mathbf{z}}_v$ according to \eqref{eq:sub2}.
	    \ENDIF
	    \STATE $\hat{\pmb{\mathbb{Z}}}[i,j] \leftarrow \phi(\hat{\mathbf{z}}_l,\hat{\mathbf{z}}_v)$
	    \STATE $\hat{\pmb{\mathbb{Z}}}[j,i] \leftarrow \phi(\hat{\mathbf{z}}_l,\hat{\mathbf{z}}_v)$
	    \ENDFOR
	    \ENDFOR
	    \ENSURE Return $\hat{\pmb{\mathbb{Z}}}$
	\end{algorithmic}
\end{algorithm}

\section{Predicting Colorectal Cancer Survival}

We begin this section with an in-depth disclosure of the data employed in our experiments, subsequently outlining the technicalities involved in undertaking these experiments. Finally, we provide a discussion of the results elicited from performing these experiments by comparing the average predicted survival curve of each method against the average actual survival curve, leveraging a devised measure, referred to as \textit{area between the curves} (ABC), discussed further on in this section.

\subsection{Colorectal Cancer Survival Data for the State of Iowa}

Our data were provided by the Iowa Cancer Registry (ICR), State Health Registry of Iowa (SHRI), and the Iowa Department of Public Health (IDPH). Each instance represents a patient who has been diagnosed with colorectal cancer and whose residence at the time of diagnosis is in the state of Iowa. The dataset consists of $n=46116$ patients and, initially, $m=71$ features. After removing identifiers and features having a large number of instances with missing values (\% missing $>$ 50\%), we were left with $m=26$ distinct features (including unprocessed geographic coordinates). After binarizing discrete features, $m=386$ (excluding geographic features). When using SBR geographical re-representation, $m=1364$ ($386$ non geographic features and $p=978$ binarized geographic features), and $m=386 + k$ when using the RR-SA geographic representation (where $k$ is parameterized and therefore user-dependent). When the Kaplan-Meier re-representation is applied to the dataset, we obtain $\by^{(i)}$ vectors having $T=53$ elements, where each element represents the patient's current vital status (alive$=1$ or dead$=0$), or a probability of survival when an instance becomes censored, as described by \eqref{eq:rerep}. Each $\tilde{t} \in \{1,\dots,53\}$ represents six months.

\begin{figure}[h]
    \centering
    \includegraphics[scale=0.045]{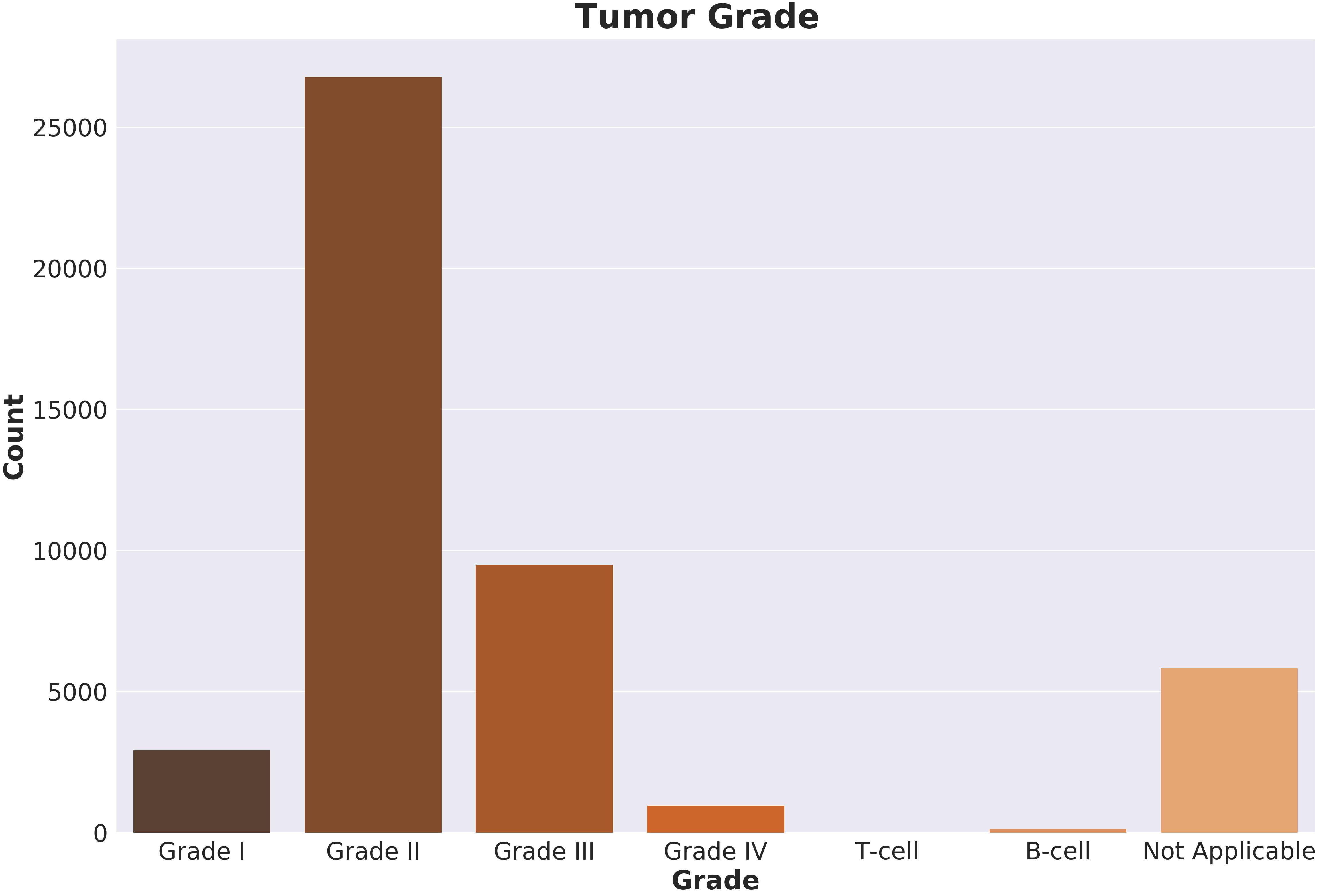}
    \caption{Tumor grade at diagnosis for patients in the state of Iowa: Years 1989 to 2013.\label{fig:grade}}
\end{figure}

\begin{figure}[h]
    \centering
    \includegraphics[scale=0.045]{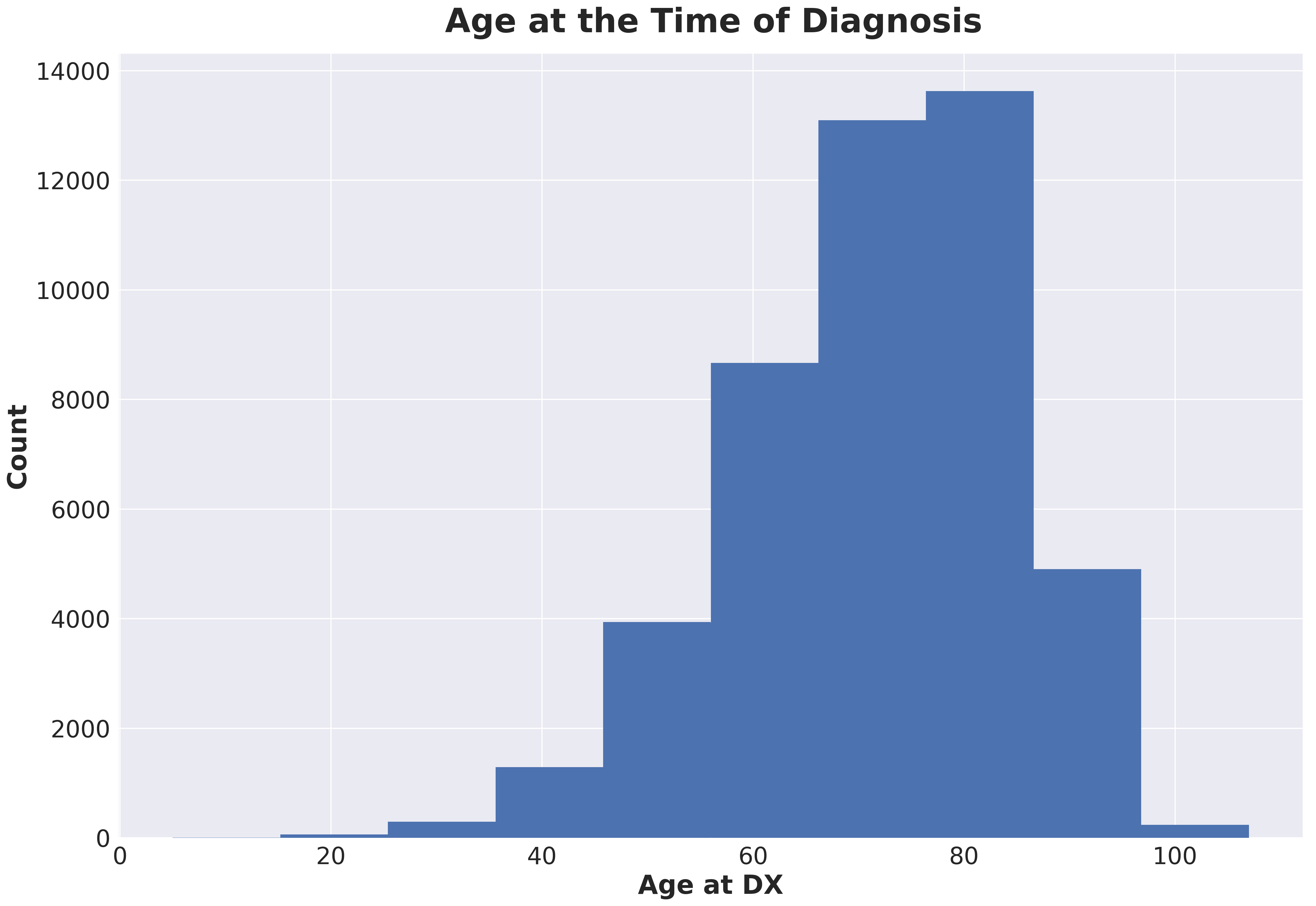}
    \caption{Age of colorectal cancer diagnosis for patients in the state of Iowa: Years 1989 to 2013.\label{fig:age}}
\end{figure}

The $24$ distinct non-geographic features pertain to various patient-specific characteristics, which can be categorized as \textit{disease-based} and \textit{demographic-based}. Disease-based features include tumor grade, tumor histology and tumor marker; we show a histogram of tumor grade in Figure \ref{fig:grade}. Demographic-based features include marital status, race, and age at diagnosis; we show a histogram of age at diagnosis in Figure \ref{fig:age}. These selected features (age and tumor grade) have been shown to be indicative of not receiving timely cancer treatment \cite{ward2013does}, which we believe will help in predicting cancer survival, although analysis of such factors is beyond the scope of this work.

\subsection{Geographically-descriptive Features for the State of Iowa}

We obtain geographically-descriptive features for the state of Iowa at the ZCTA-level of spatial granularity from the US Census Bureau's American FactFinder 2 website. Three different geographically-descriptive features were obtained for each of the 978 ZCTAS in Iowa: population age demographics, land type, and median household income. Population age demographics and land type are categorical features and were represented in terms of proportional bins (e.g., ``percentage of population aged 0-5 years'').

\subsection{Predictive Setting, Paramaterization and Results}

As outlined in the introduction, we wish to address the following:
\begin{enumerate}
    \item On average, can colorectal cancer survival curves be
reasonably predicted for patients in the state of Iowa?
    \item Do geographic features improve the quality of predicted
colorectal cancer survival curves for patients in the state
of Iowa?
    \item Do richer geographical feature representations improve predictive
performance more than simpler representations?
    \item Can predictive performance be further improved by altering the RR-SA procedure to accommodate adjacency-descriptive geographical feature pairings (i.e., RR-SSA)?
    \item Which RR-SSA representation improves predictive performance the most: binary (bin) or full?
\end{enumerate}

To such an end, we propose to use $10$-fold validation where, for each fold, we find a $\mathtt{g}^*$ for each of the following types of model:
\begin{enumerate}[label=(\roman*)]
\item A model constructed using no geographical features (No Geo).
\item A model constructed using SBR-derived geographical features, as outlined by Figure \ref{fig:sbr-arch} (SBR).
\item Models constructed using RR-SA-derived geographical features, as outlined by Figure \ref{fig:rr-sc-arch}, where the values $k=10,20,30,40$ will be explored (RR-SA).
\item Models construct using RR-SSA-derived geographical features, as outlined by Figure \ref{fig:rr-ssa-arch}, using the \textit{binary-based} adjacency representation, where the values $k=10,20,30,40$ will be explored (RR-SSA (bin)).
\item Models construct using RR-SSA-derived geographical features, as outlined by Figure \ref{fig:rr-ssa-arch}, using the \textit{full} adjacency representation, where the values $k=10,20,30,40$ will be explored (RR-SSA (bin)).
\end{enumerate}

We then assess predictive performance by computing each model's average survival curve prediction on the test set, taken over $10$ folds, as compared to the actual average survival curve, taken over all $\by^{(i)}$, using a measure termed \textit{area between curves} (ABC) that measures the area-wise disparity between the actual and predicted curves \cite{lash2017learning}.

\subsubsection{Model Parameterization}

Our models are constructed using Tensorflow, employing fully connected layers, trained using sigmoidal cross entropy as the loss function $\mathcal{L}(\cdot)$. The logistic activation function is used for all nodes. Each model is trained using a maximum of $2500$ epochs with batch size ranging from $5\%$ to $20\%$. While the connectedness of the architecture, activation function, epochs, and batch size are all tunable parameters, we elect to focus on finding the optimal number of hidden layers and corresponding hidden nodes for each layer (note that epochs of $1000$, $1500$, $2000$, and $2500$ were explored). Table \ref{tab:params}, in the Appendix section, shows the average optimal architecture for each of the models, taken over the 10 folds.

\begin{figure*}[!htp]
    \centering
    \begin{subfigure}[]{.32\linewidth}
        \centering
        \captionsetup{justification=centering}
        \includegraphics[scale=.25]{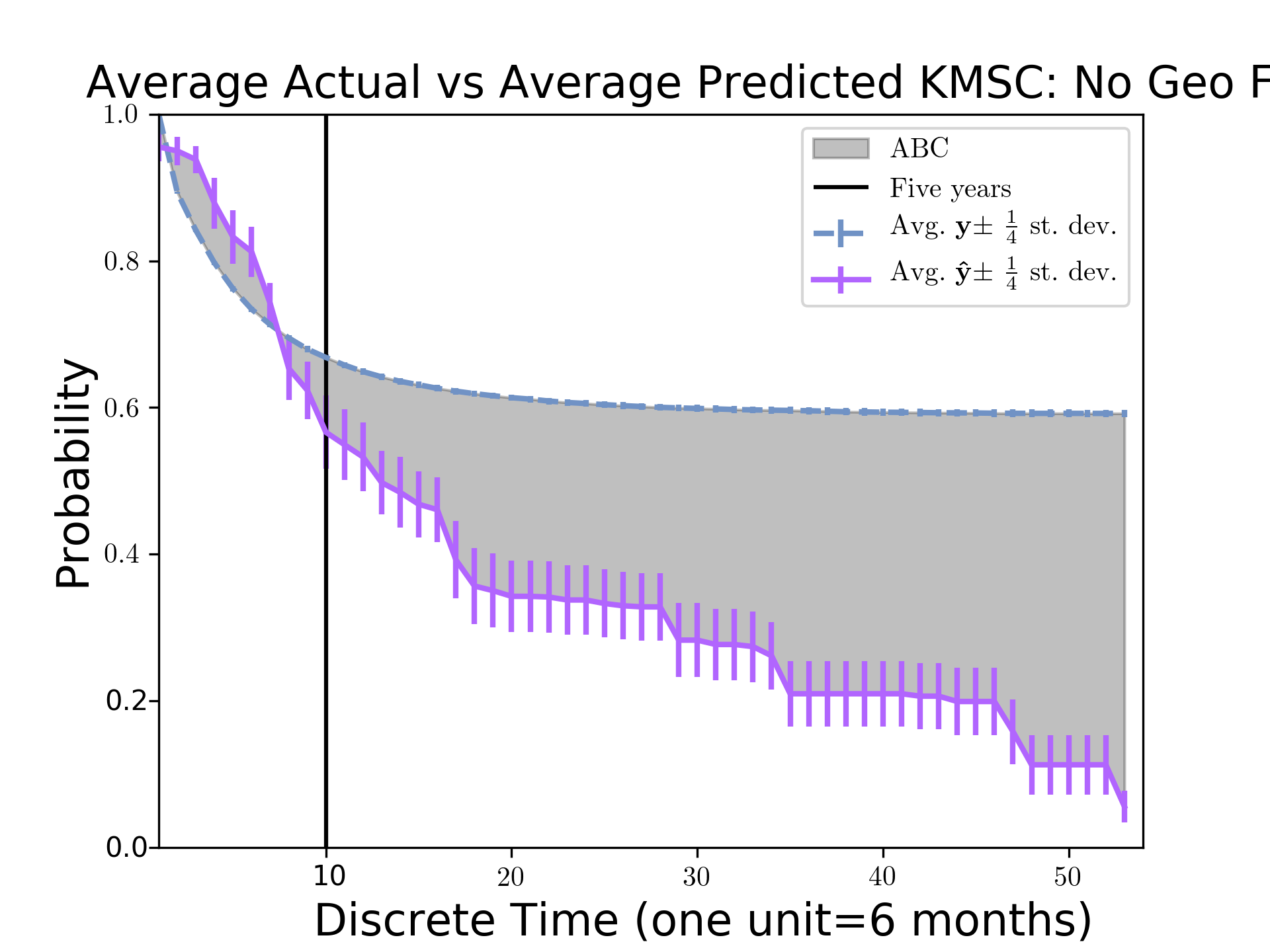}
        \caption{No geo feats\\(ABC=14.32).\label{fig:scnogeo}}
    \end{subfigure}
    \begin{subfigure}[]{.32\linewidth}
        \centering
        \captionsetup{justification=centering}
        \includegraphics[scale=.25]{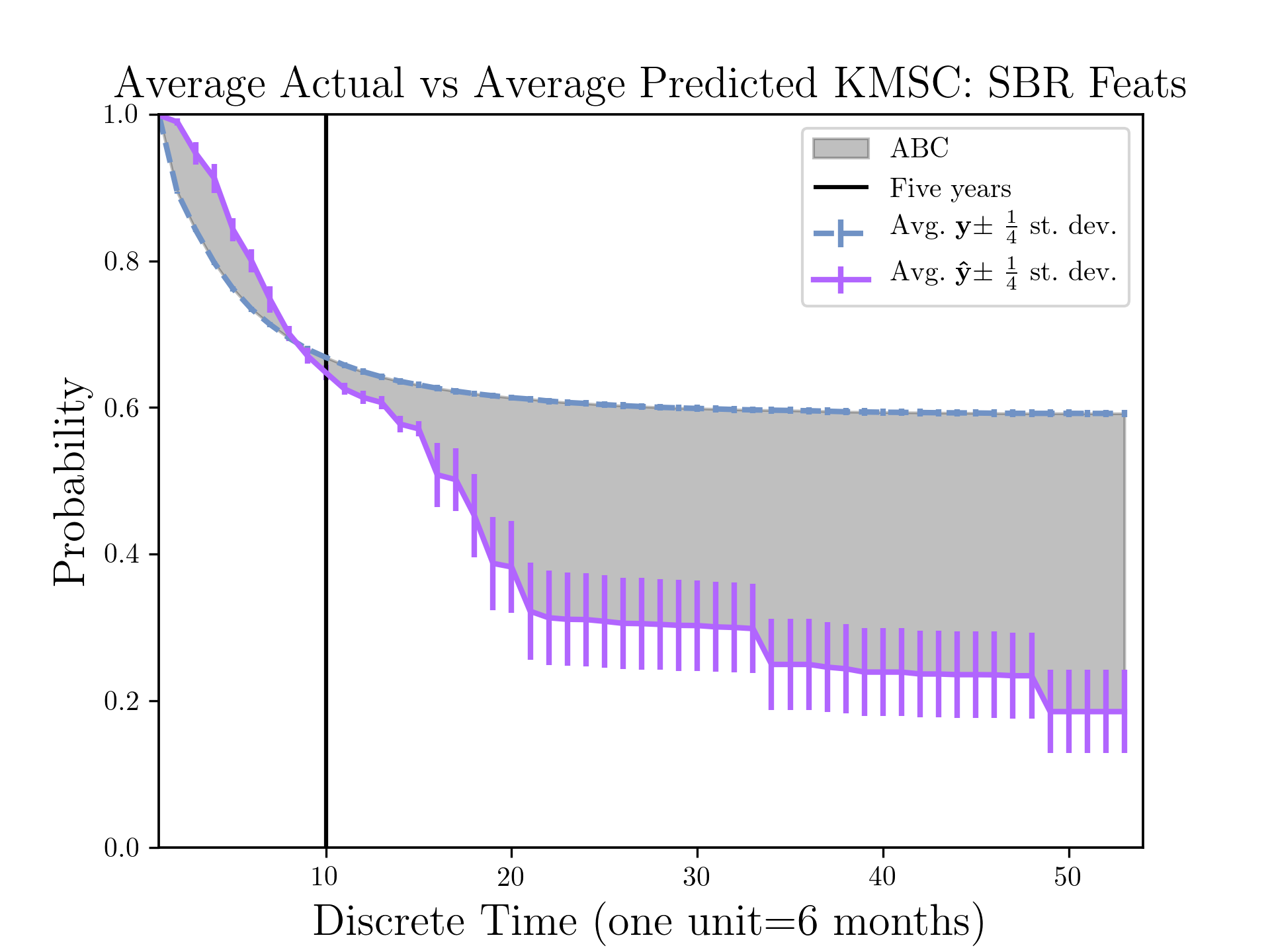}
        \caption{SBR\\(ABC=12.60).\label{fig:scwithgeo}}
    \end{subfigure}\par
    \begin{subfigure}[]{.32\linewidth}
        \centering
        \captionsetup{justification=centering}
        \includegraphics[scale=.25]{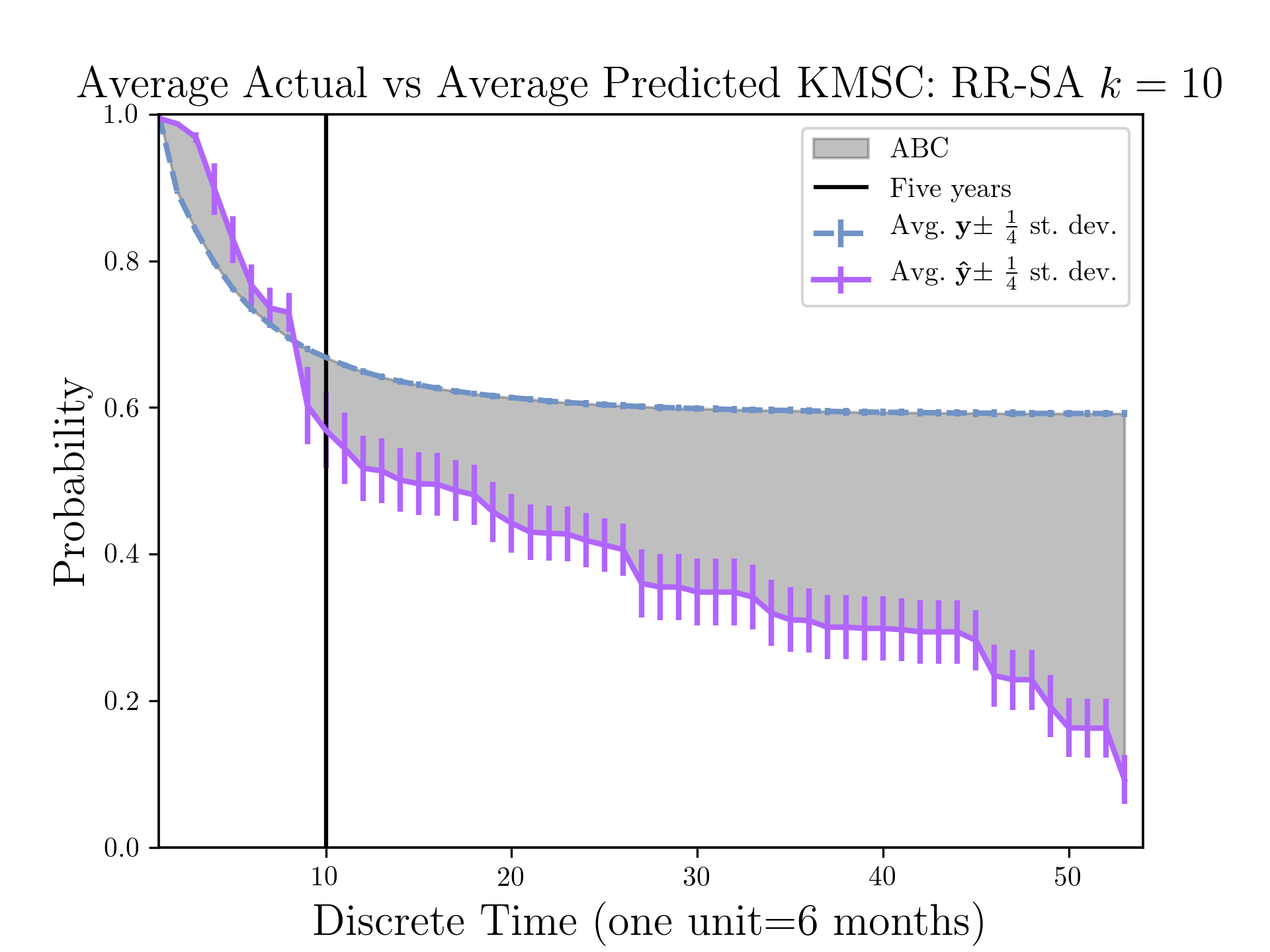}
        \caption{RR-SA, $k=10$\\(ABC=11.41).\label{fig:sc10geo}}
    \end{subfigure}
    \begin{subfigure}[]{.32\linewidth}
        \centering
        \captionsetup{justification=centering}
        \includegraphics[scale=.25]{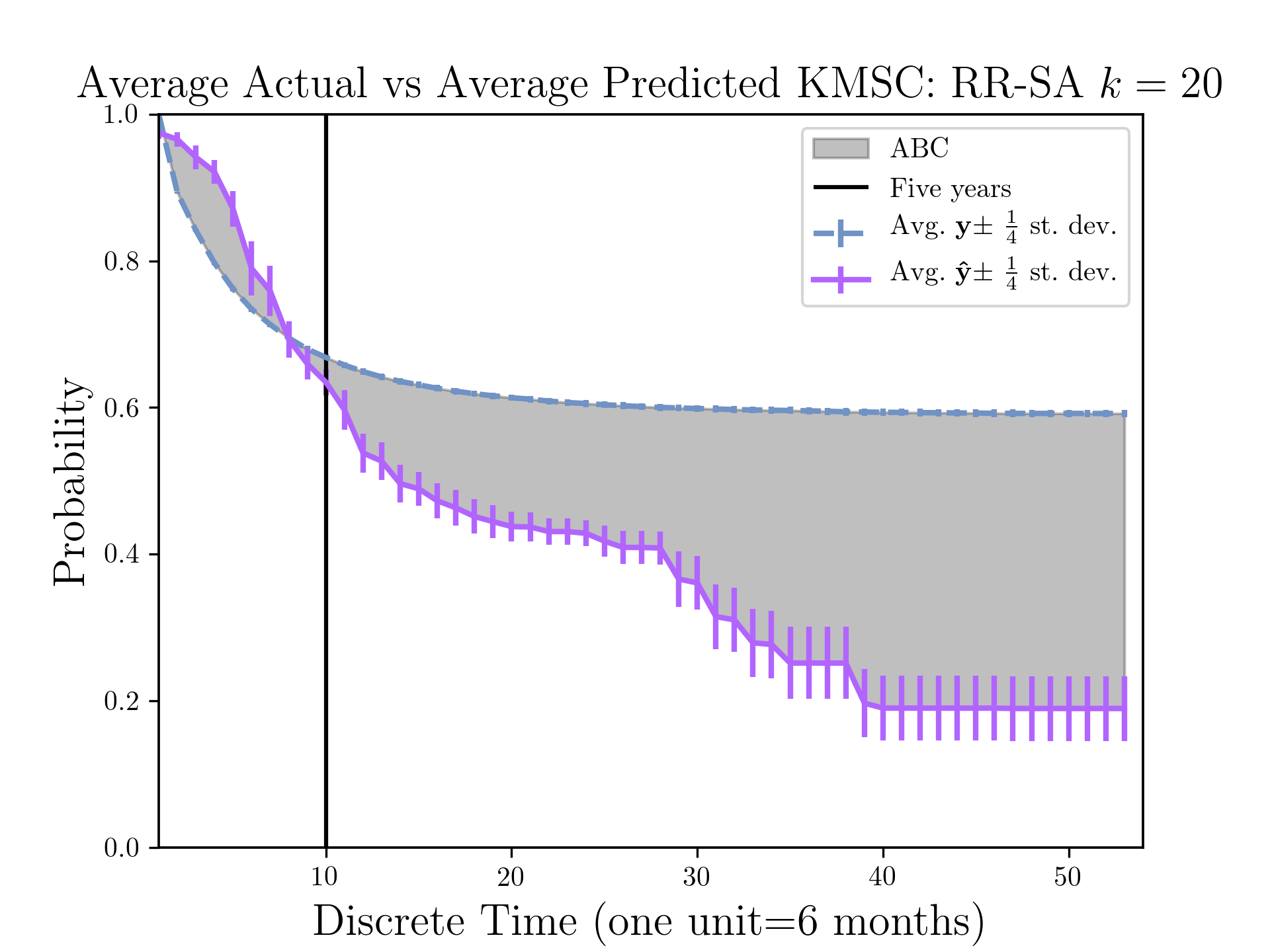}
        \caption{RR-SA, $k=20$\\(ABC=12.31).\label{fig:sc20geo}}
        \label{}
    \end{subfigure}
    \begin{subfigure}[]{.32\linewidth}
        \centering
        \captionsetup{justification=centering}
        \includegraphics[scale=.25]{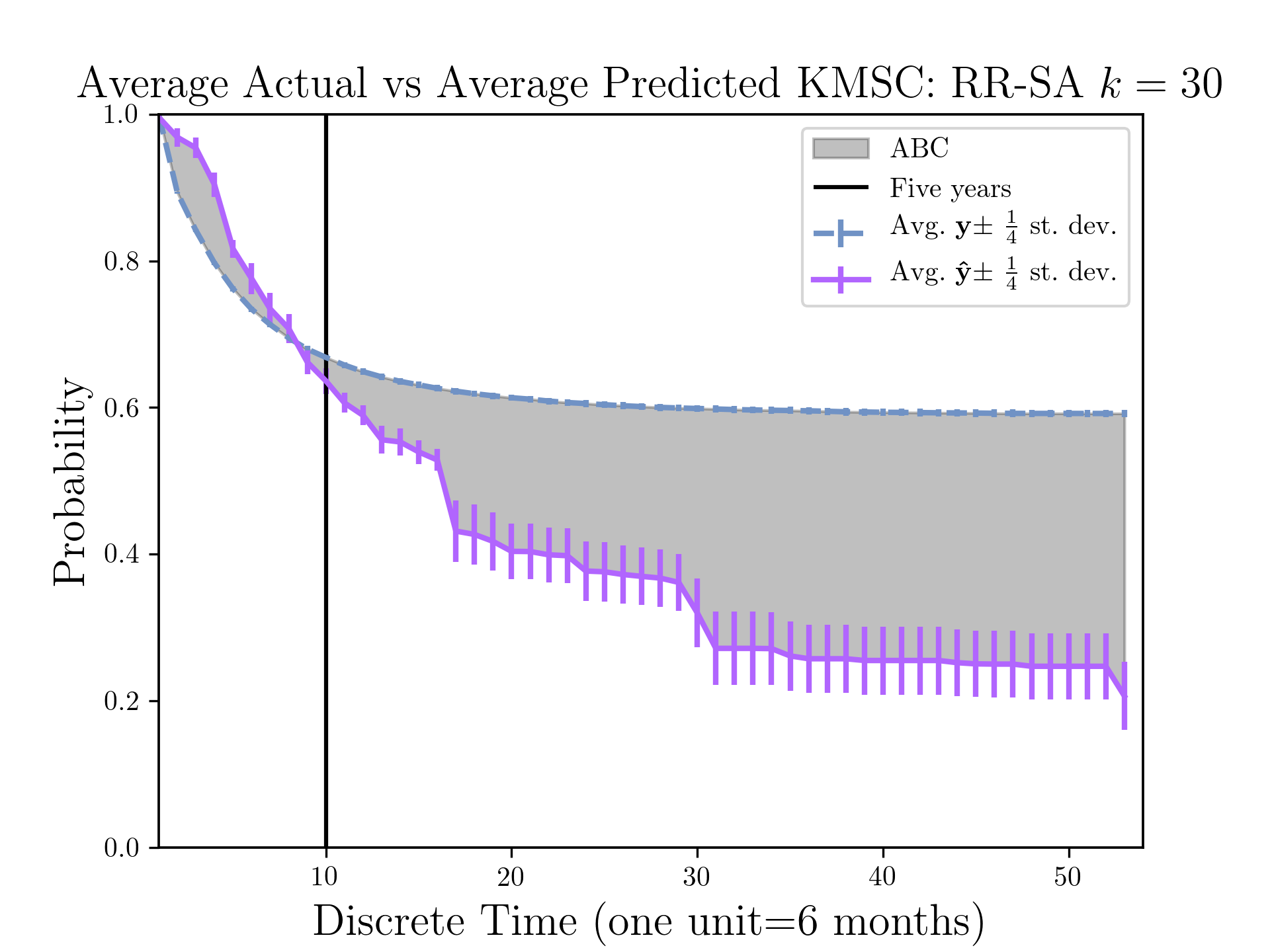}
        \caption{RR-SA, $k=30$\\(ABC=11.65).\label{fig:sc30geo}}
    \end{subfigure}\par
    \begin{subfigure}[]{.32\linewidth}
        \centering
        \captionsetup{justification=centering}
        \includegraphics[scale=.25]{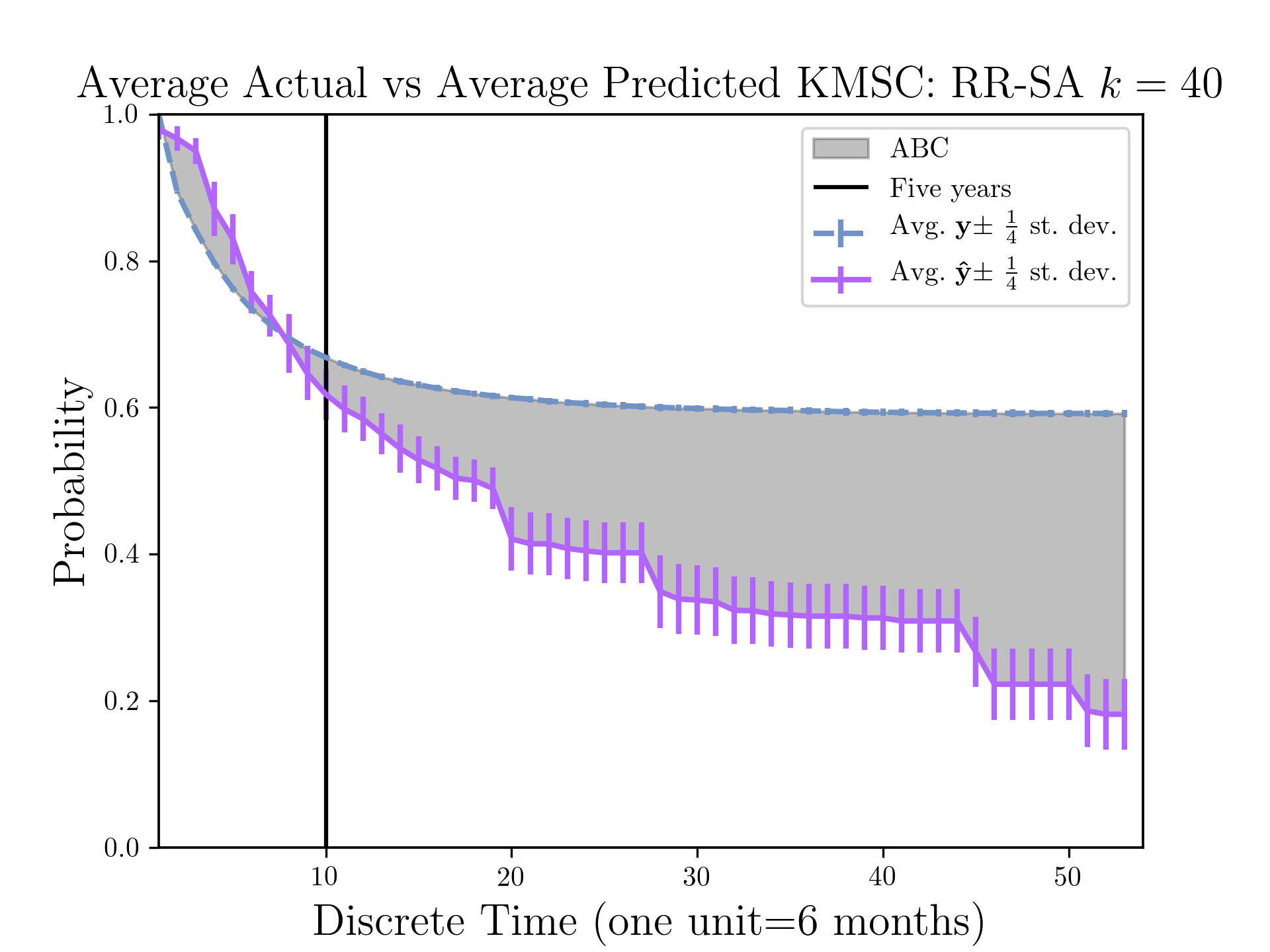}
        \caption{RR-SA, $k=40$\\(ABC=10.77).\label{fig:sc40geo}}
    \end{subfigure}
    \begin{subfigure}[]{.32\linewidth}
        \centering
        \captionsetup{justification=centering}
        \includegraphics[scale=.25]{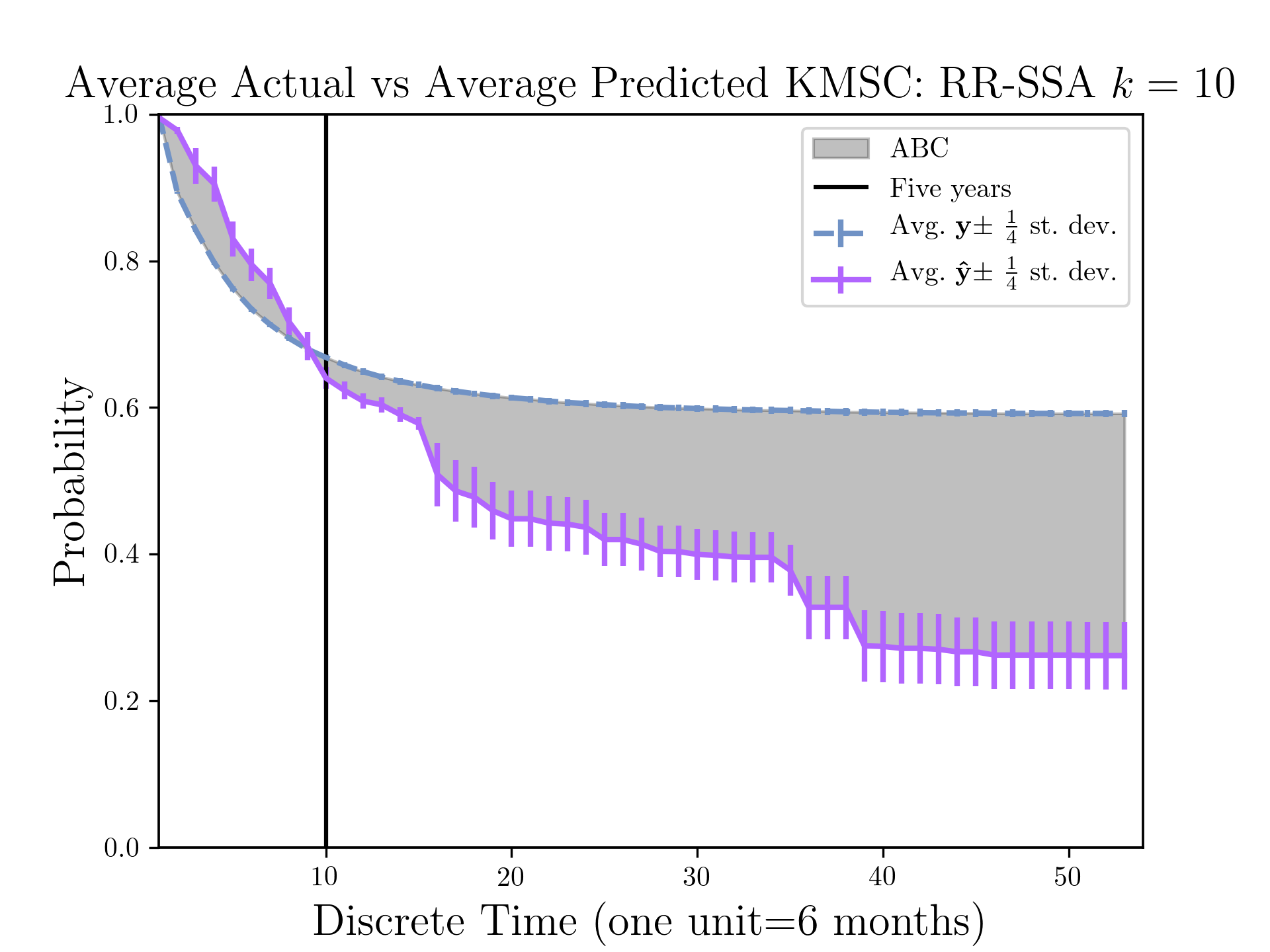}
        \caption{RR-SSA (bin), $k=10$\\(ABC=9.805).\label{fig:ssabin10geo}}
    \end{subfigure}
    \begin{subfigure}[]{.32\linewidth}
        \centering
        \captionsetup{justification=centering}
        \includegraphics[scale=.25]{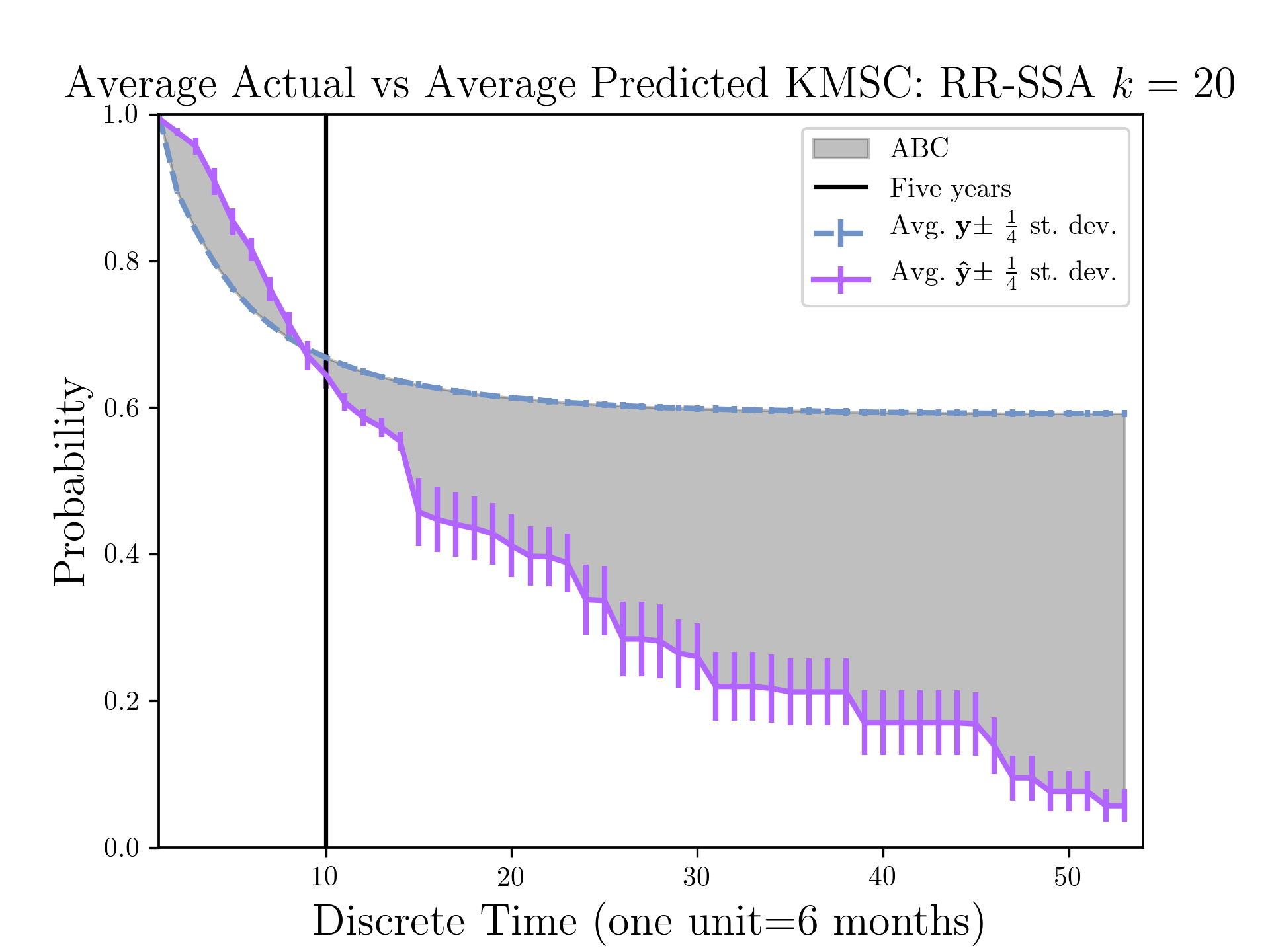}
        \caption{RR-SSA (bin), $k=20$\\(ABC=14.555).\label{fig:ssabin20geo}}
    \end{subfigure}\par
    \begin{subfigure}[]{.32\linewidth}
        \centering
        \captionsetup{justification=centering}
        \includegraphics[scale=.25]{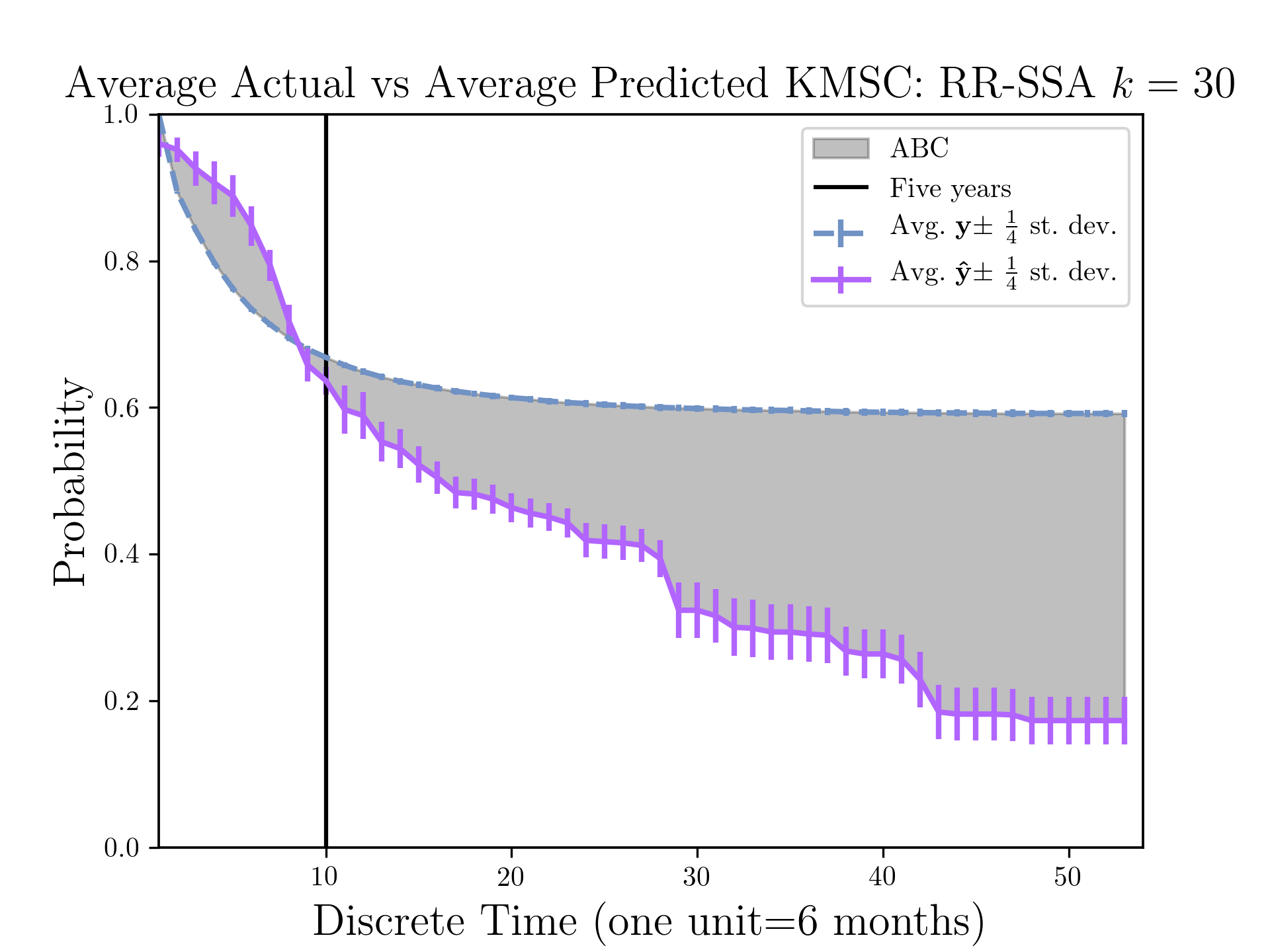}
        \caption{RR-SSA (bin), $k=30$\\(ABC=11.850).\label{fig:ssabin30geo}}
    \end{subfigure}
    \begin{subfigure}[]{.32\linewidth}
        \centering
        \captionsetup{justification=centering}
        \includegraphics[scale=.25]{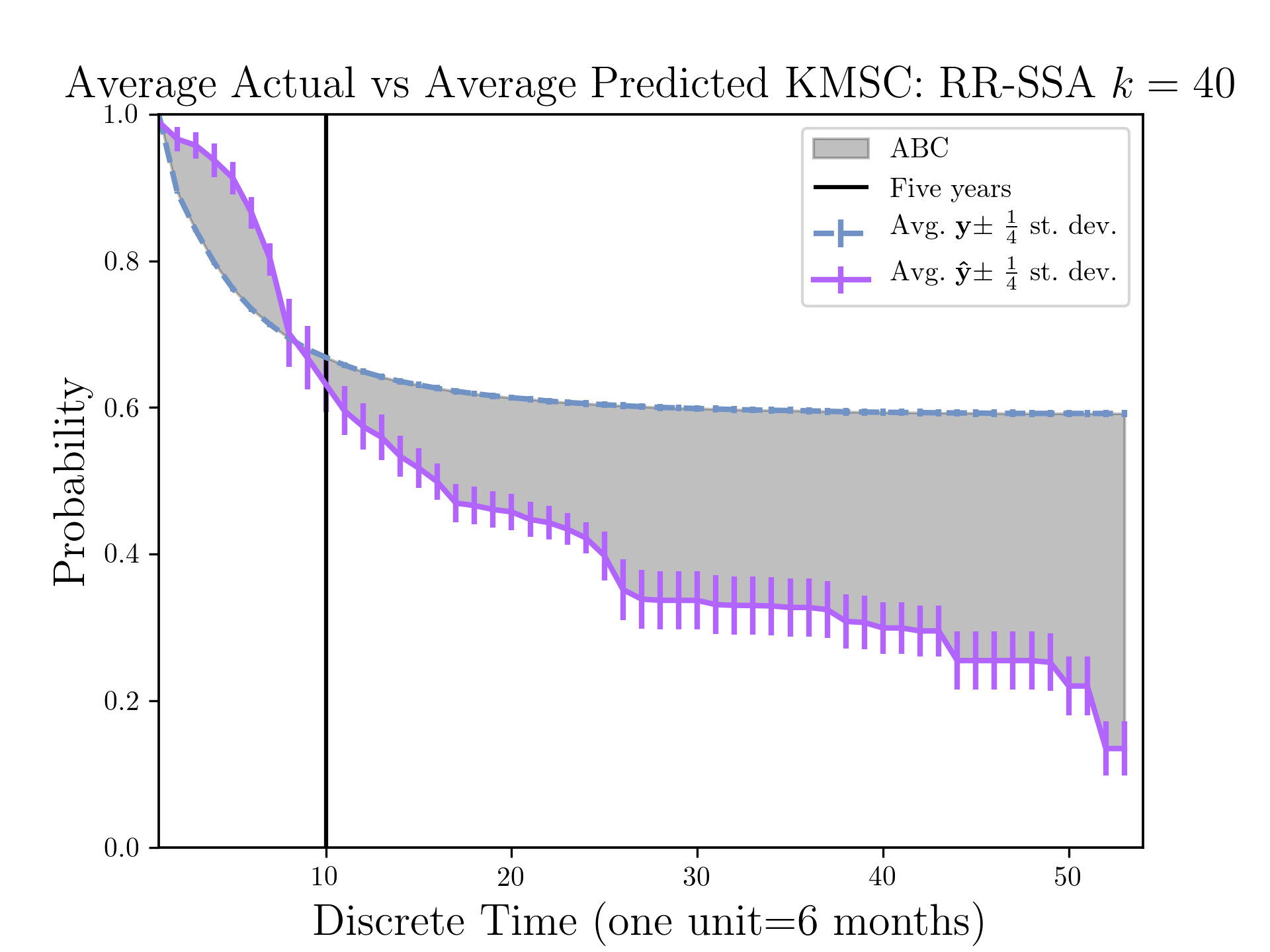}
        \caption{RR-SSA (bin), $k=40$\\(ABC=11.205).\label{fig:ssabin40geo}}
    \end{subfigure}
    \begin{subfigure}[]{.32\linewidth}
        \centering
        \captionsetup{justification=centering}
        \includegraphics[scale=.25]{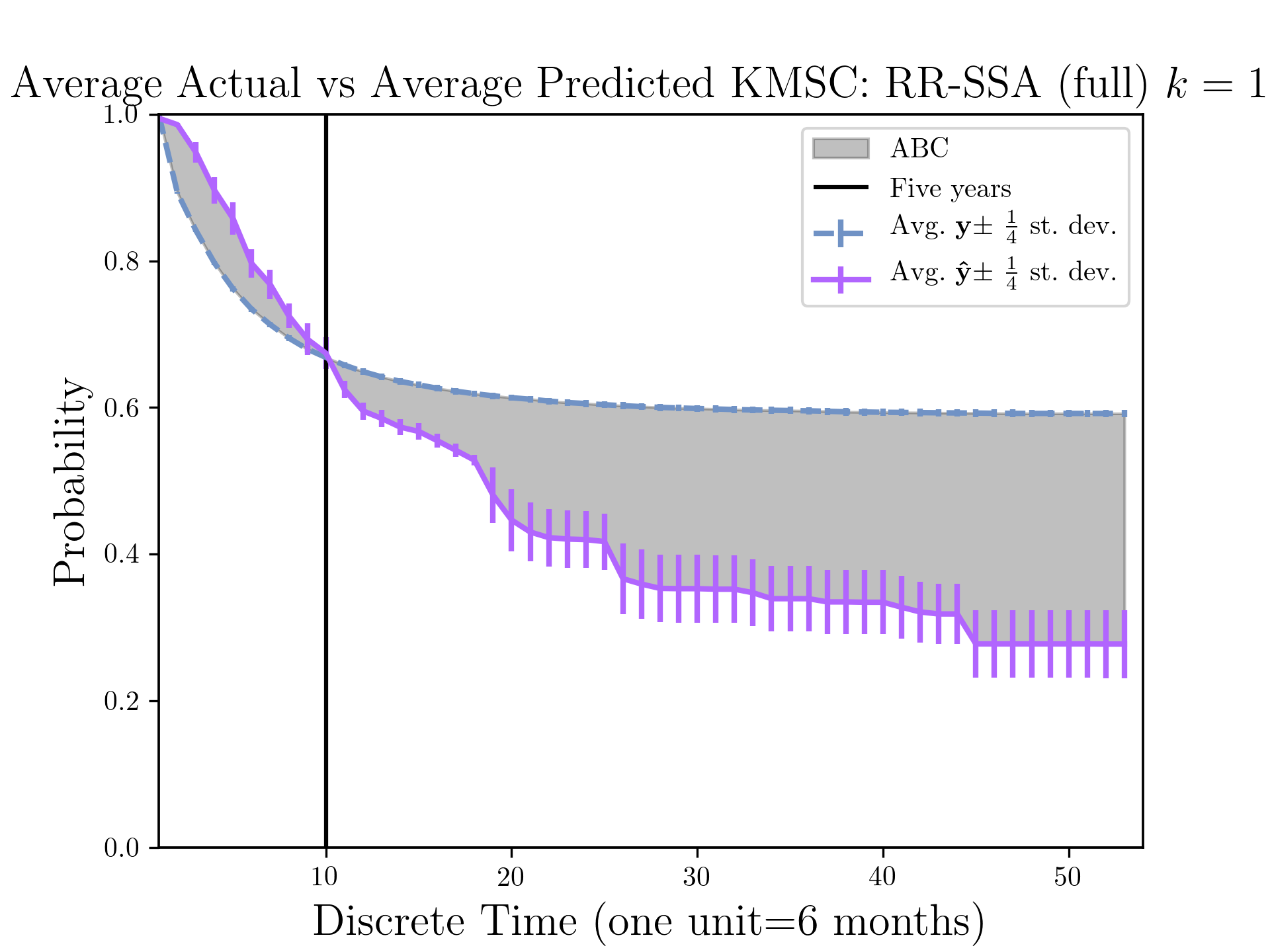}
        \caption{RR-SSA (full), $k=10$\\(ABC=9.820).\label{fig:ssafull10geo}}
    \end{subfigure}\par
    \begin{subfigure}[]{.32\linewidth}
        \centering
        \captionsetup{justification=centering}
        \includegraphics[scale=.25]{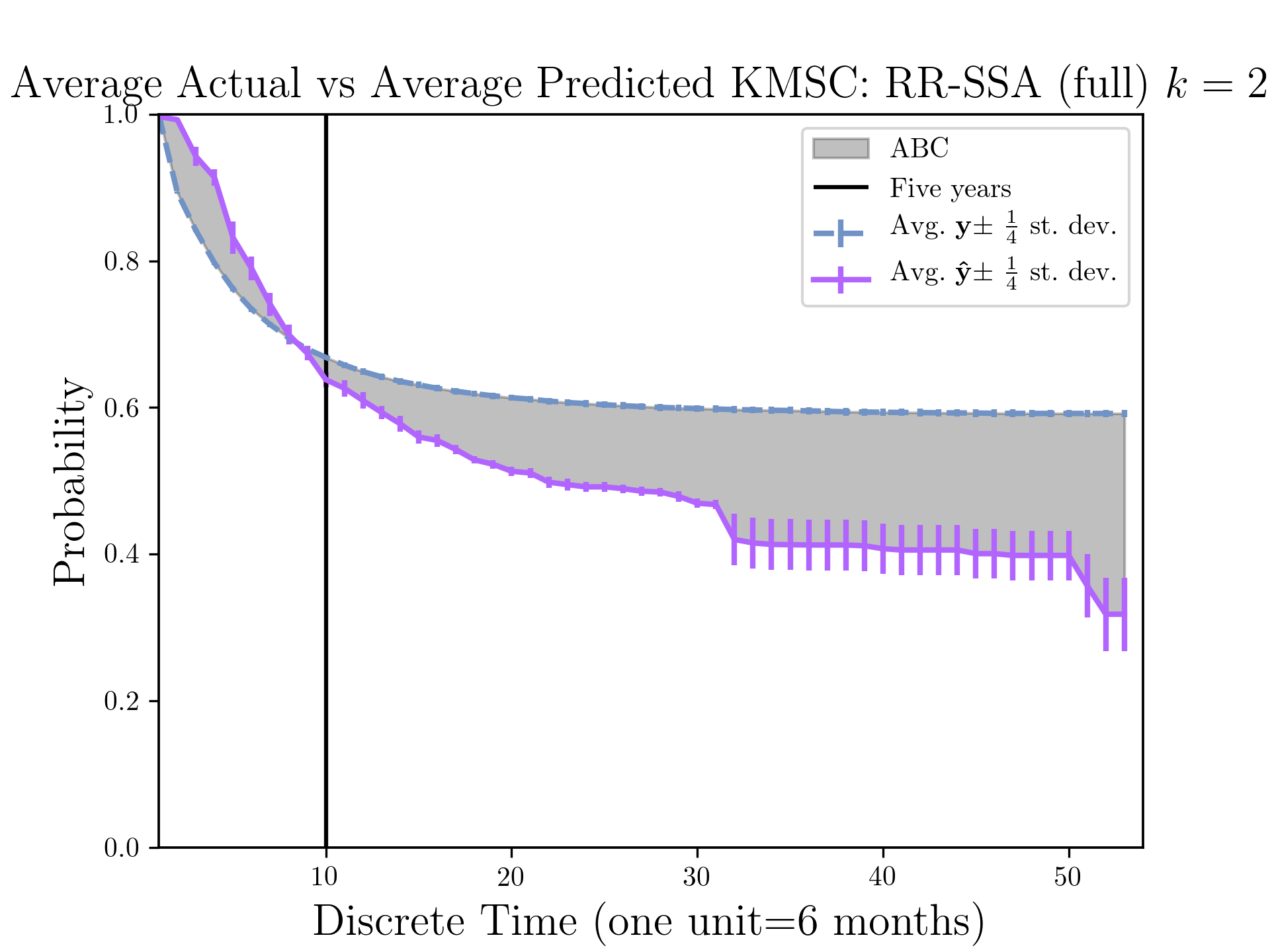}
        \caption{RR-SSA (full), $k=20$\\(ABC=6.657).\label{fig:ssafull20geo}}
    \end{subfigure}
    \begin{subfigure}[]{.32\linewidth}
        \centering
        \captionsetup{justification=centering}
        \includegraphics[scale=.25]{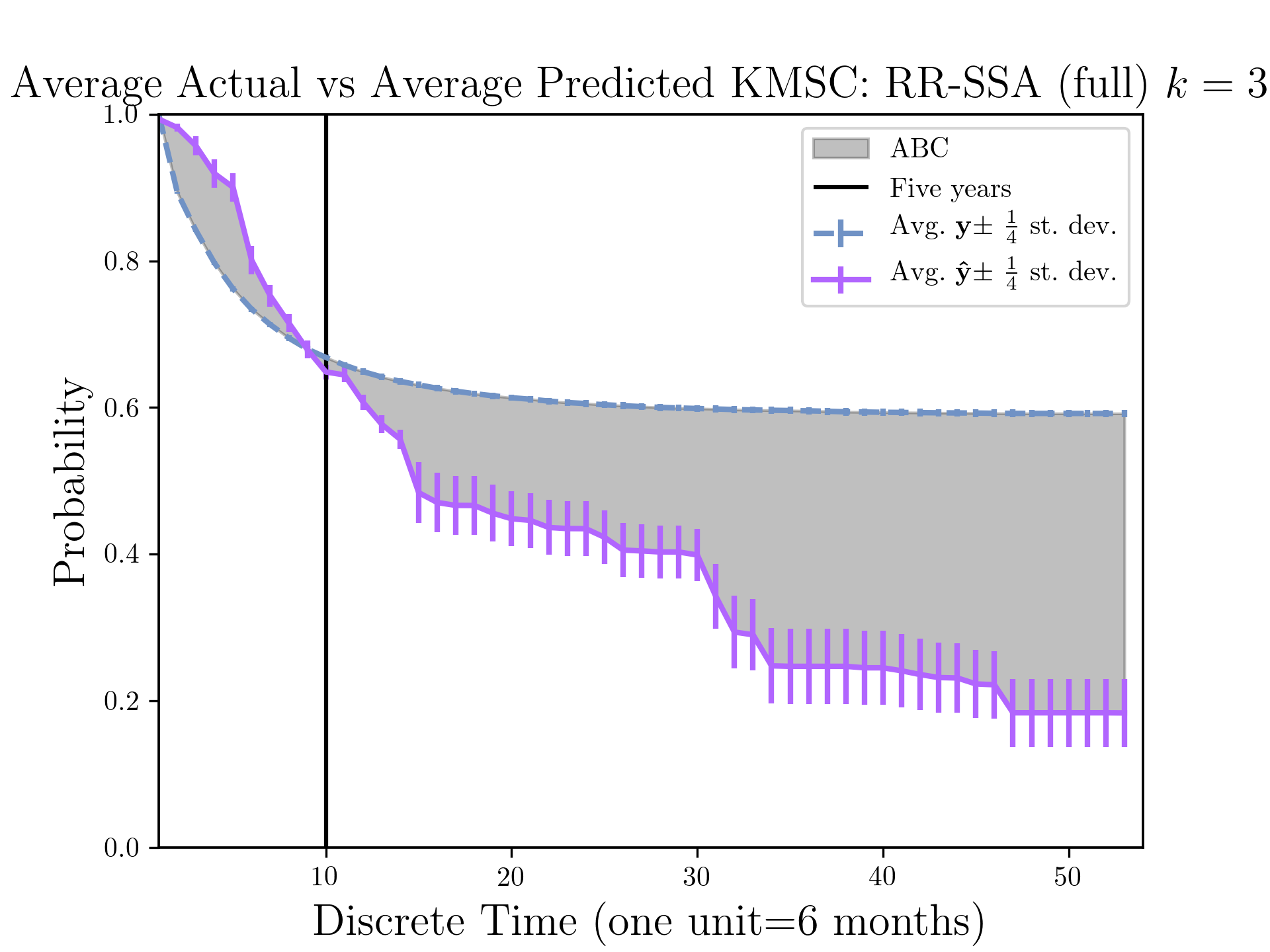}
        \caption{RR-SSA (full), $k=30$\\(ABC=11.724).\label{fig:ssafull30geo}}
    \end{subfigure}
    \begin{subfigure}[]{.32\linewidth}
        \centering
        \captionsetup{justification=centering}
        \includegraphics[scale=.25]{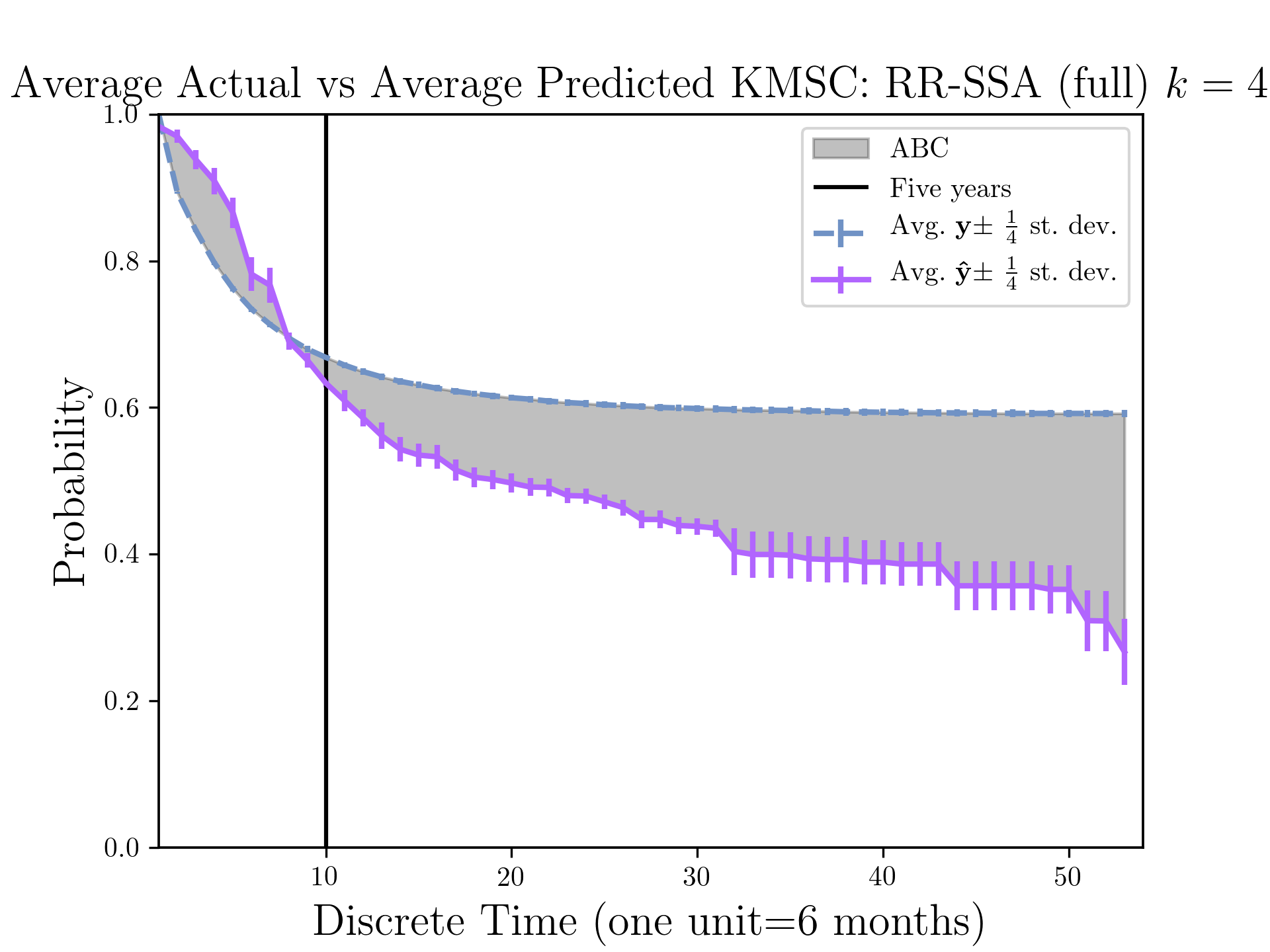}
        \caption{RR-SSA (full), $k=40$\\(ABC=7.818).\label{fig:ssafull40geo}}
    \end{subfigure}
    \caption{Actual vs. Predicted; $k$ (when specified) denotes the parameterized $k$ for spectral analysis, and ABC represents the \textit{area between curves}.\label{fig:avpsc}}
\end{figure*}

\subsubsection{Average Actual vs Average Predicted Survival}

The results comparing the average actual survival curve against the average predicted survival curve, by model, are presented in Figure \ref{fig:avpsc}. Henceforth, these curves will simply be referred to as \textit{actual} and \textit{predicted}. In these figures we also shade the region between the actual and predicted curves and provide a value representing the total area covered by this region. We will use this measure, developed in Lash et al.,~2017 \cite{lash2017learning}, referred to as \textit{area between the curves} (ABC for short), as a means of comparing the predictive quality of the 14 different models (where lower ABC is better).

Comparing Figure \ref{fig:scnogeo} with Figures \ref{fig:scwithgeo} through \ref{fig:ssafull40geo} we first see that the addition of geographical features has uniformly improved the quality of the predictions, on average, as can be observed visually and by comparing ABC values. That is, with the exception of RR-SSA (bin) $k=20$, which suggests that it is important to tune the spectral analysis $k$ value when using such representations.

Secondly, comparing Figure \ref{fig:scwithgeo} with Figures \ref{fig:sc10geo} through \ref{fig:sc40geo}, we observe that models using richer geographical representations (RR-SA) perform better (\ref{fig:sc10geo} - \ref{fig:sc40geo}) than a model trained using a simple representation (\ref{fig:scwithgeo}). Furthermore, employing SSA-based representations yield even better improvements over SBR, depending on the parameterized value of $k$.

\begin{figure*}[!htp]
    \centering
    \begin{subfigure}[]{.24\linewidth}
        \includegraphics[scale=.028]{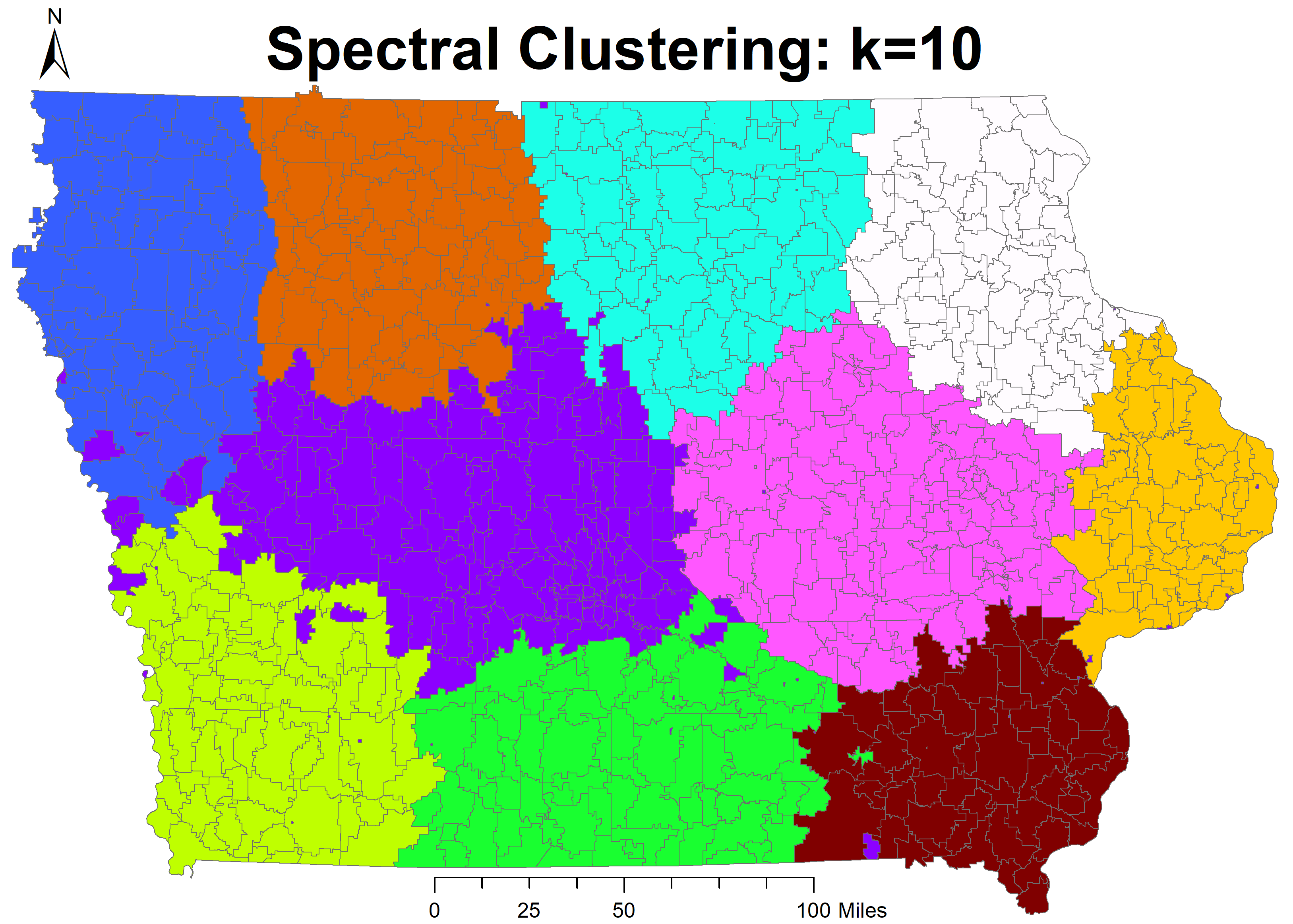}
    \end{subfigure}
    \begin{subfigure}[]{.24\linewidth}
        \includegraphics[scale=.028]{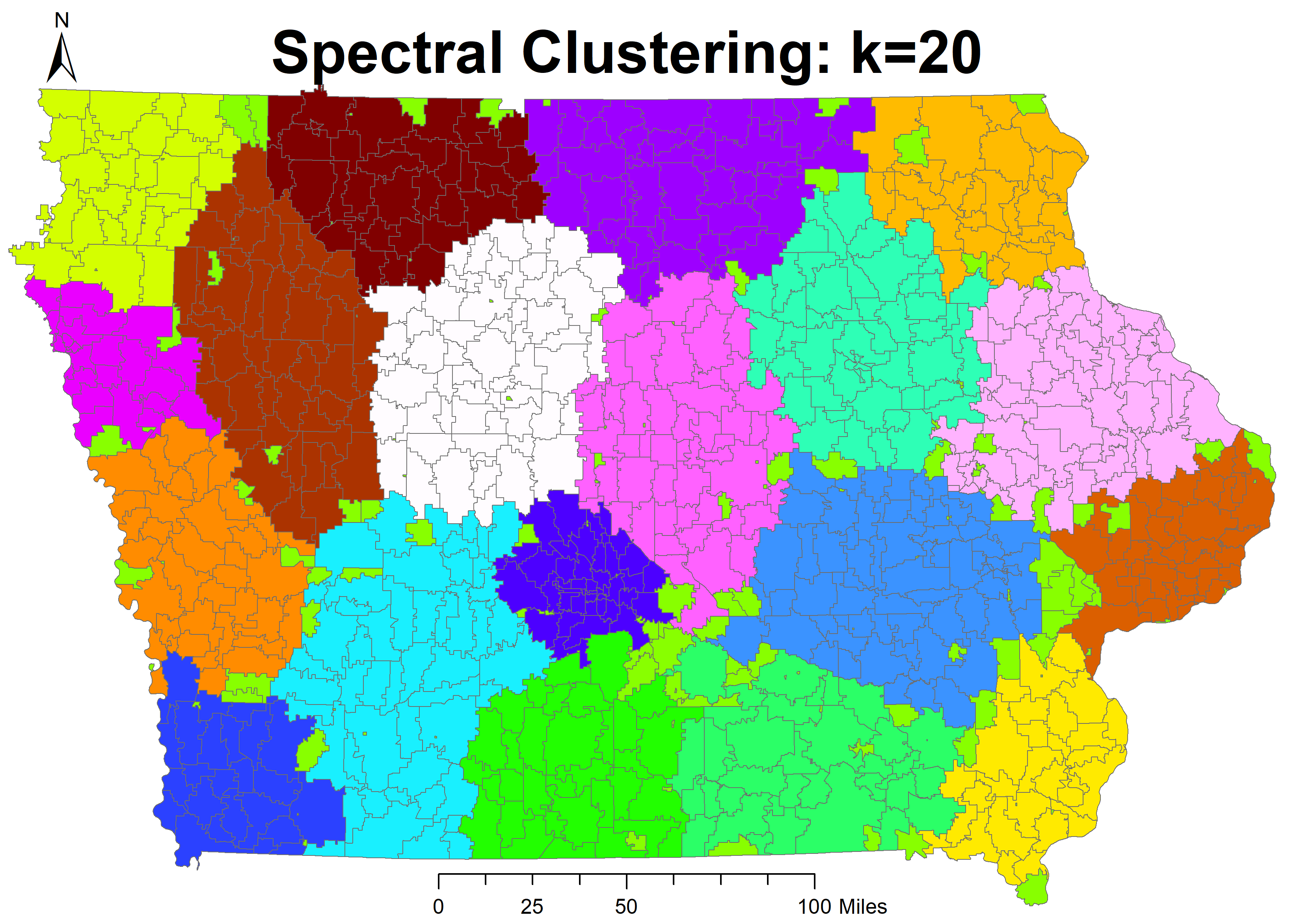}
    \end{subfigure}
    \begin{subfigure}[]{.24\linewidth}
        \includegraphics[scale=.028]{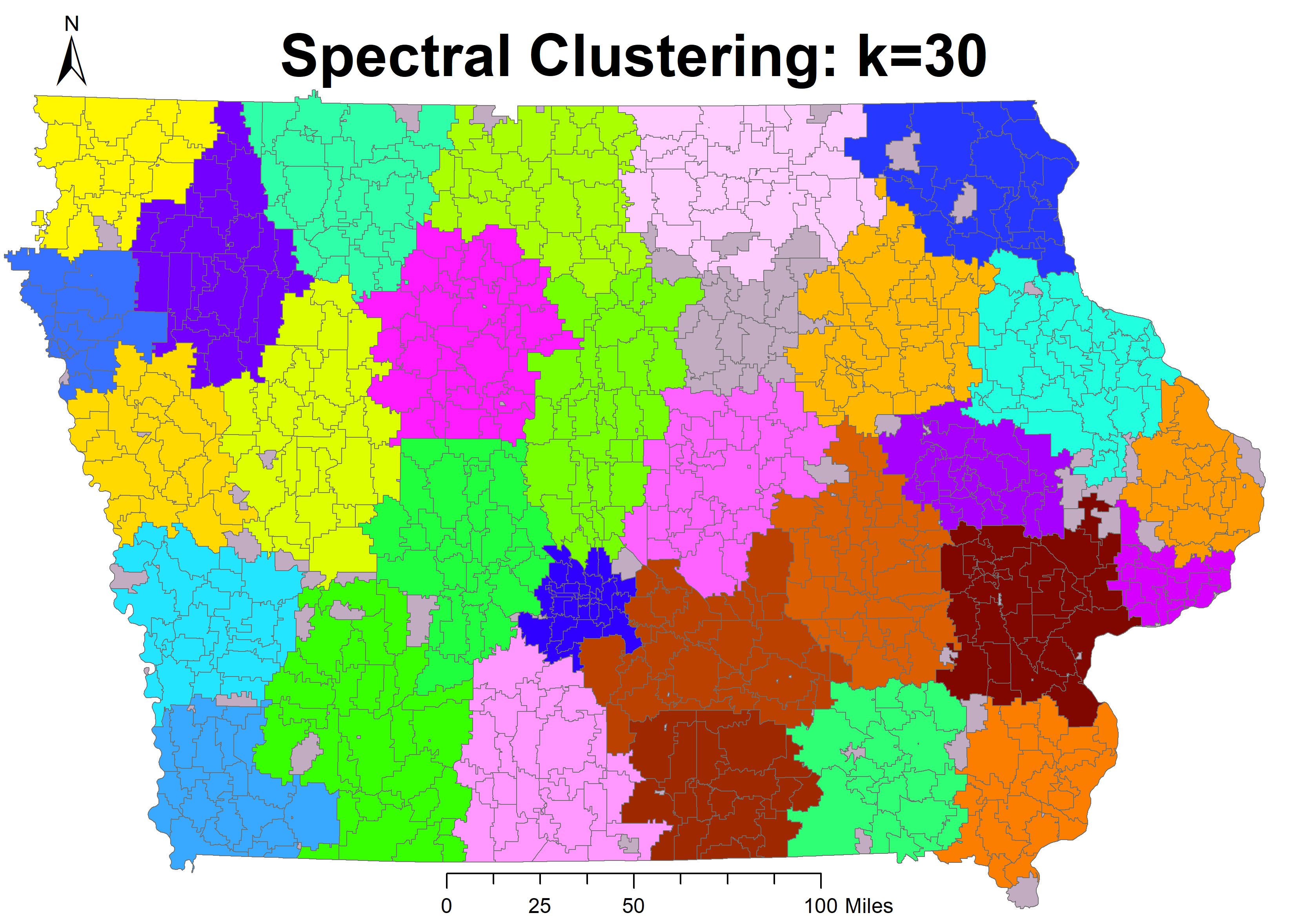}
    \end{subfigure}
    \begin{subfigure}[]{.24\linewidth}
        \includegraphics[scale=.028]{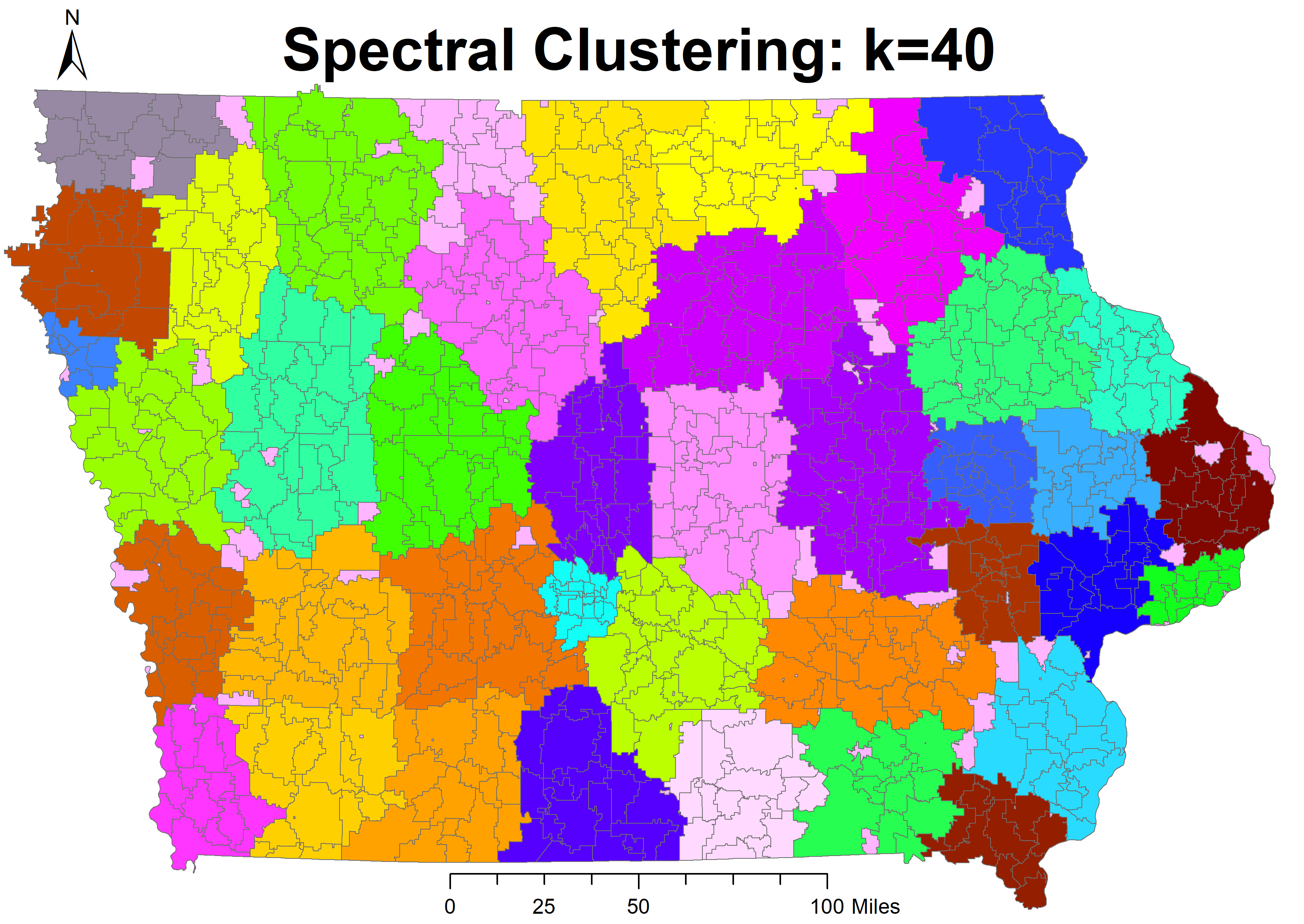}
    \end{subfigure}\par
    \begin{subfigure}[]{.24\linewidth}
        \includegraphics[scale=.115]{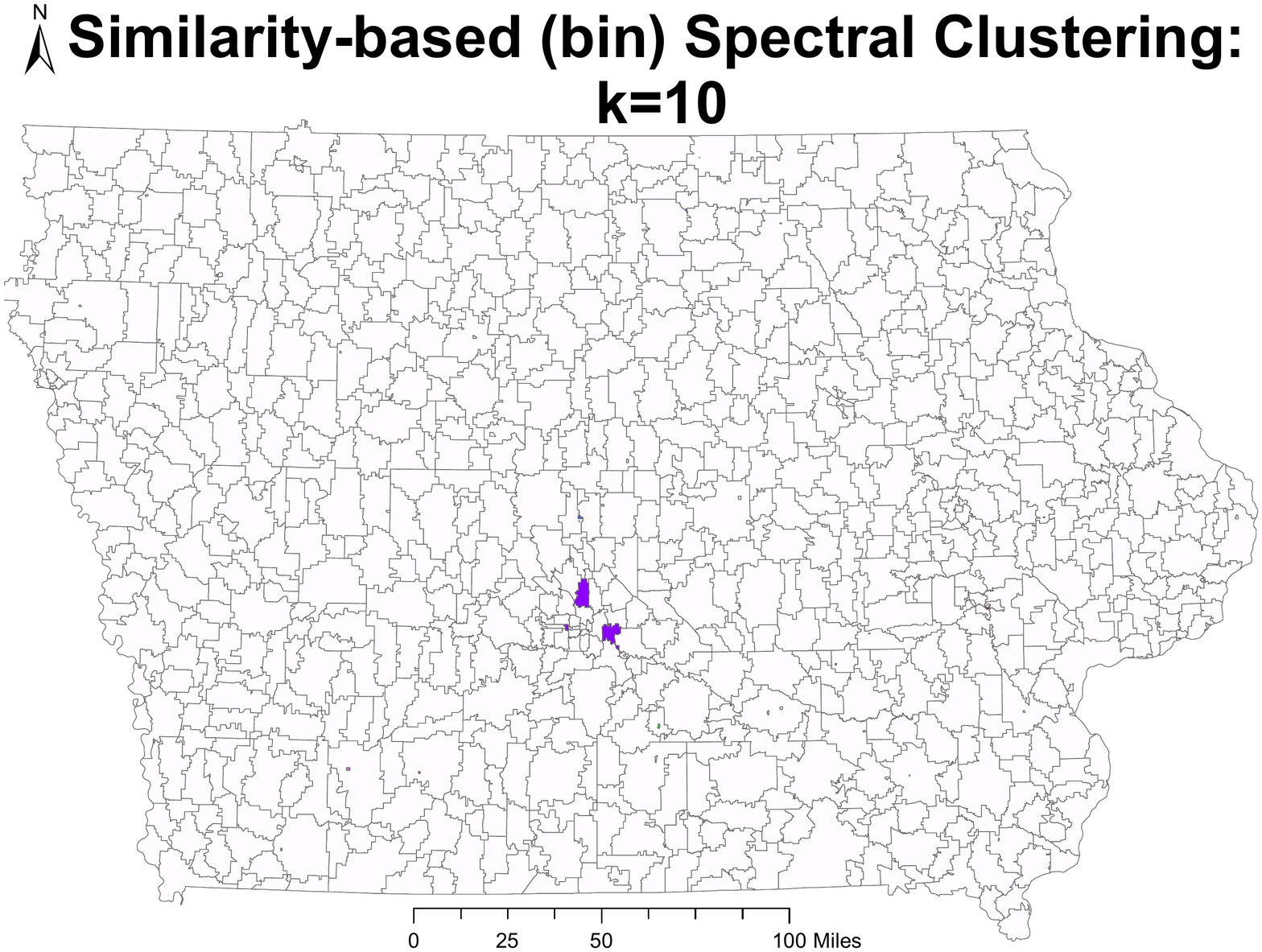}
    \end{subfigure}
    \begin{subfigure}[]{.24\linewidth}
        \includegraphics[scale=.115]{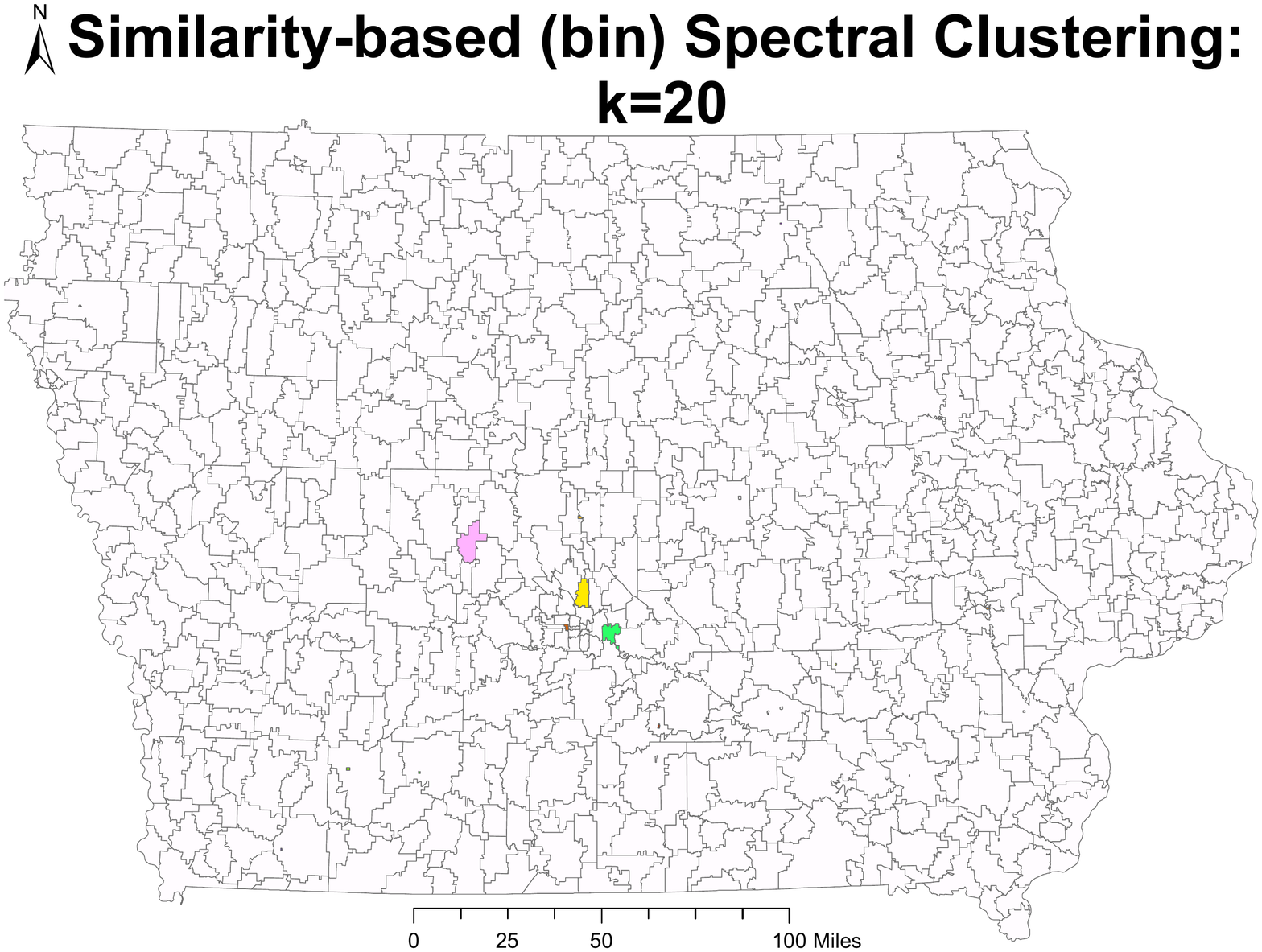}
    \end{subfigure}
    \begin{subfigure}[]{.24\linewidth}
        \includegraphics[scale=.115]{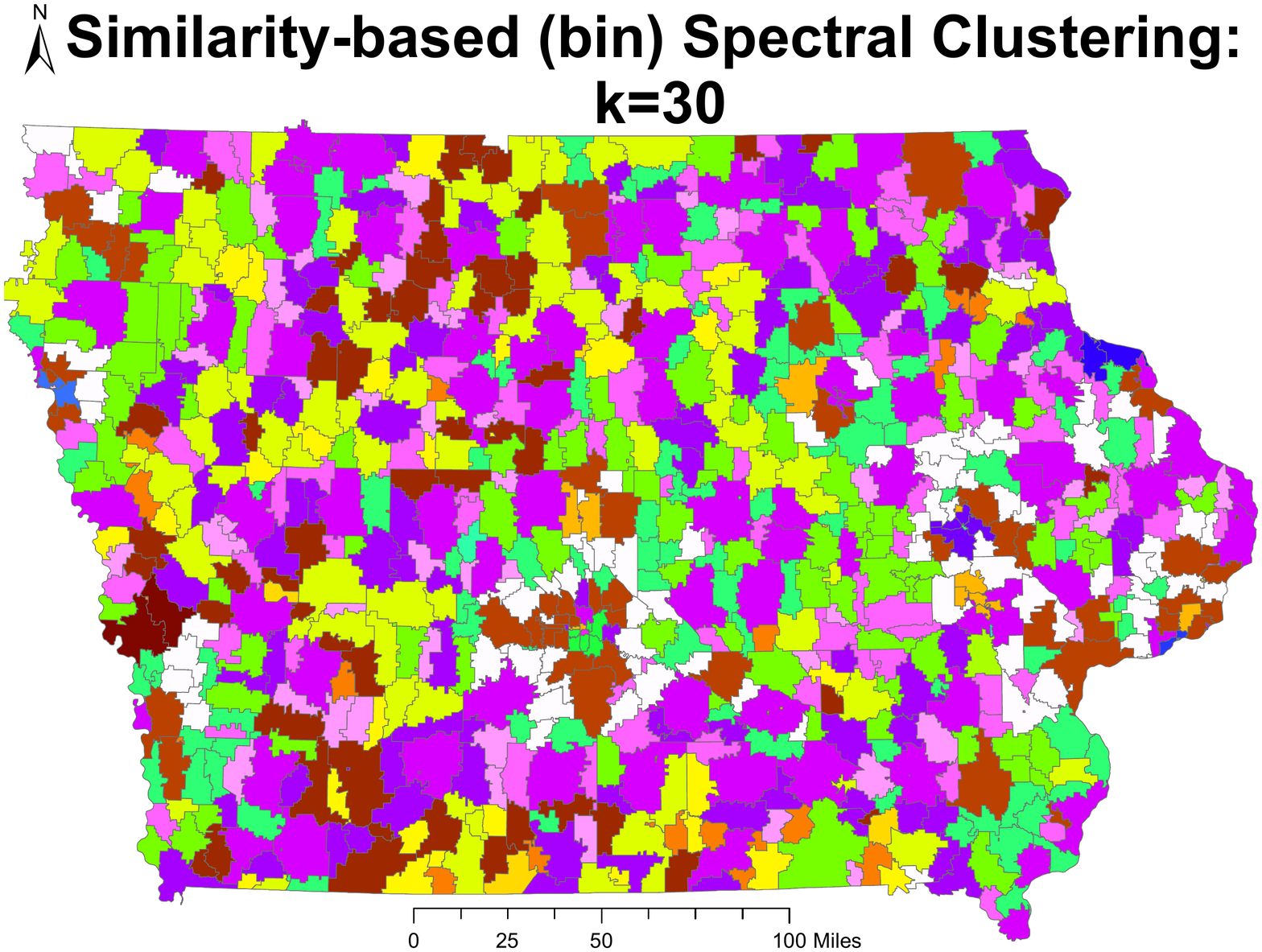}
    \end{subfigure}
    \begin{subfigure}[]{.24\linewidth}
        \includegraphics[scale=.115]{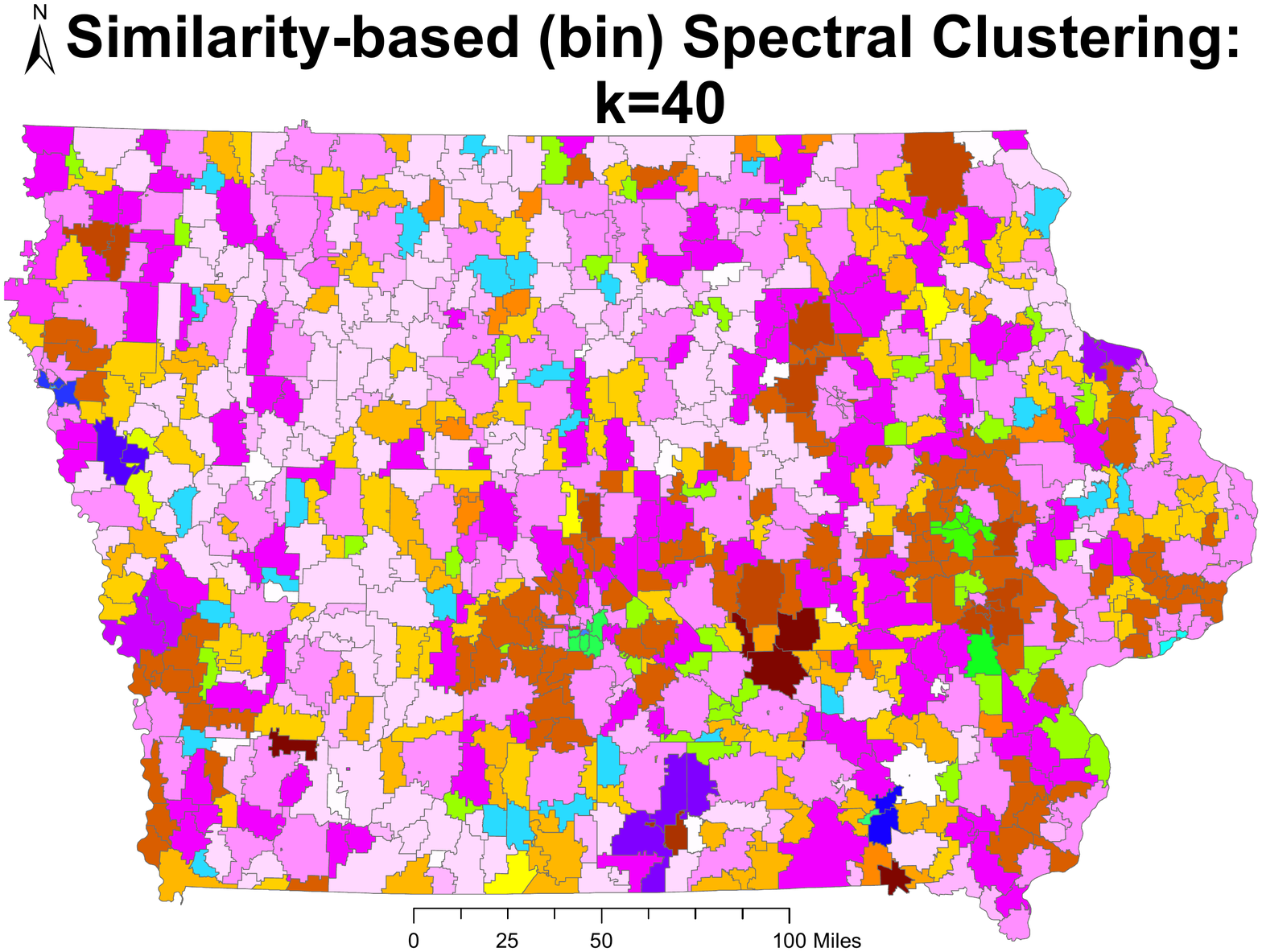}
    \end{subfigure}\par
    \begin{subfigure}[]{.24\linewidth}
        \includegraphics[scale=.115]{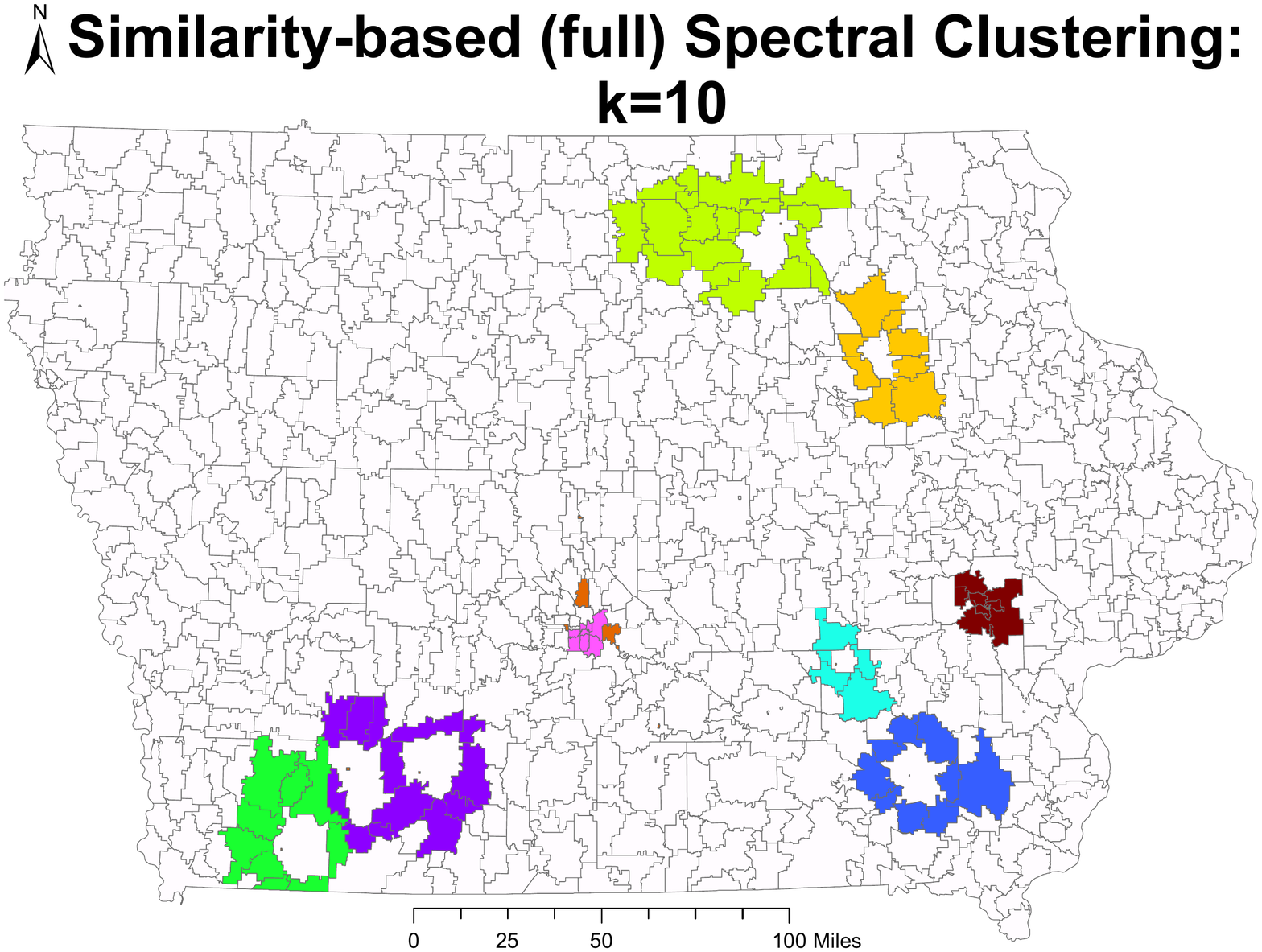}
    \end{subfigure}
    \begin{subfigure}[]{.24\linewidth}
        \includegraphics[scale=.115]{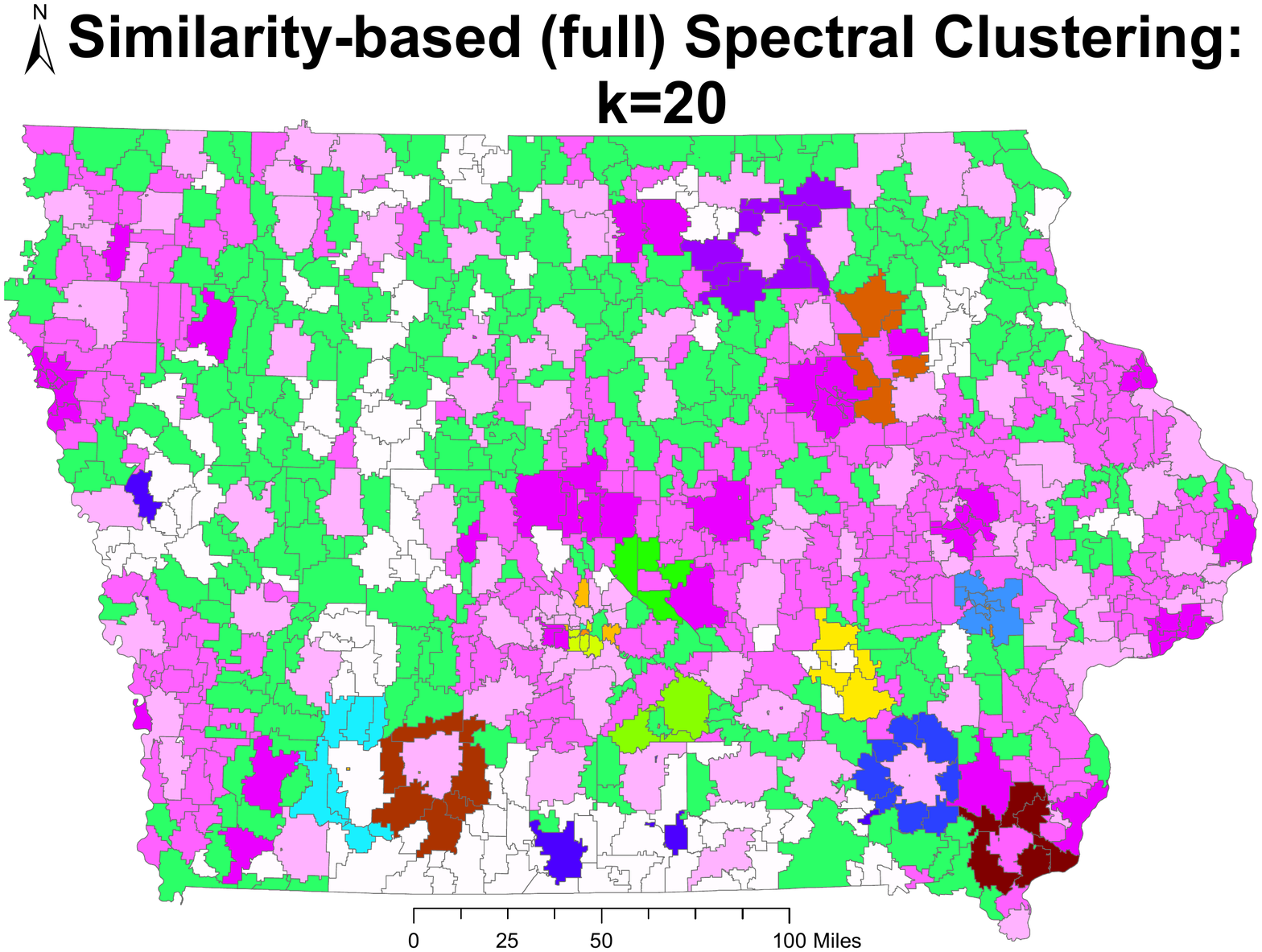}
    \end{subfigure}
    \begin{subfigure}[]{.24\linewidth}
        \includegraphics[scale=.115]{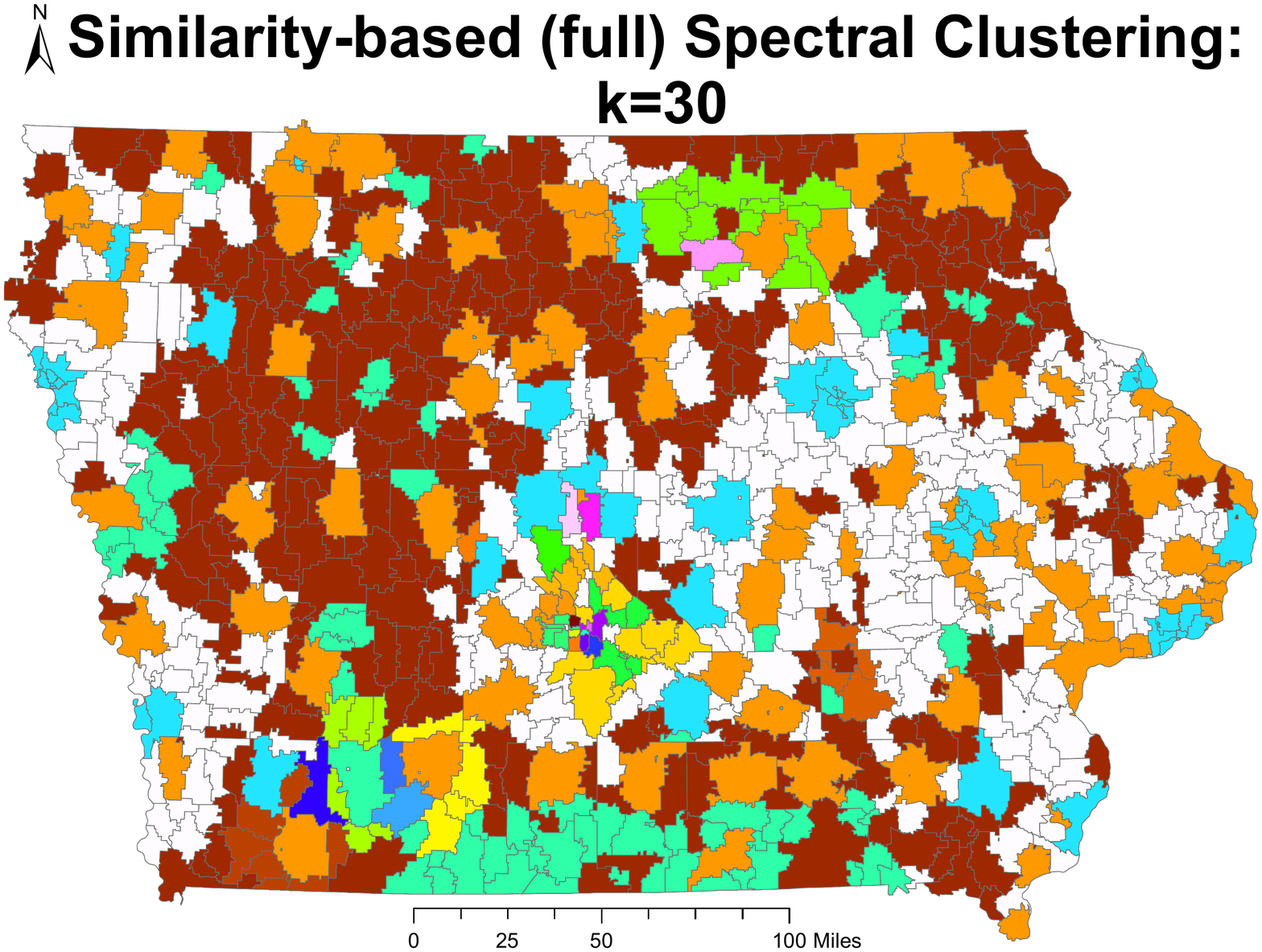}
    \end{subfigure}
    \begin{subfigure}[]{.24\linewidth}
        \includegraphics[scale=.115]{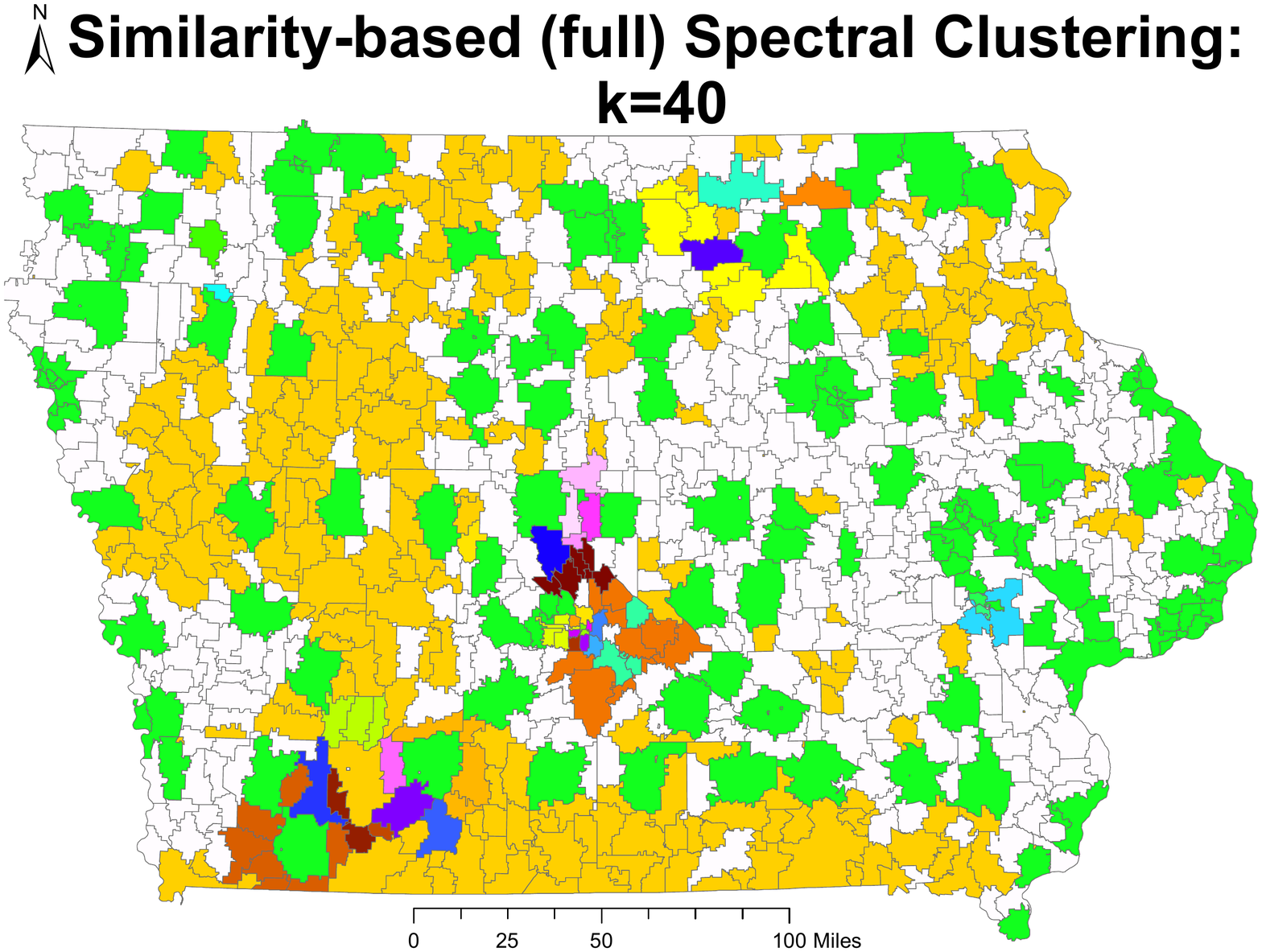}
    \end{subfigure}
    \caption{Spectral clustering results for $k=10,20,30,40$, where color denotes cluster membership. Row one represents RR-SA results, row two represents RR-SSA (bin) results, and row three represents RR-SSA (full) clustering results.}
    \label{fig:sc_map}
\end{figure*}

However, there are also RR-SA model performance differences depending on the parameterized $k$ value. Interestingly, there seems to exist a non-linear relationship between $k$ and performance, with $k=10$ outperforming $k=20$, and $k=30$ outperforming $k=10$; $k=40$ performs the best out of all models. We believe this nonlinear relationship may be accounted for by the fact that higher values of $k$ lead to more localized models, yet can also produce sparse, disjointed clusters. This point is supported by our clustering visualizations reported in Figure \ref{fig:sc_map} and discussed in Section 3.3.3. These nonlinear response observations can also be extended to RR-SSA (bin) and RR-SSA (full).

Comparing RR-SA with RR-SSA representations, we can see even greater improvement in our predictions, on average. In fact, by employing RR-SSA (full), we achieve a 38.2\% relative improvement in ABC value when comparing the best RR-SSA (full) result ($k=20$) with the best RR-SA result ($k=40$). Interestingly, and perhaps not entirely unexpectedly, RR-SSA (bin) obtained less predictive improvement when compared with RR-SSA (full), but is able to improve upon the RR-SA result.

Curiously, however, depending upon the parameterized $k$ value, RR-SSA (bin) performs worse than models induced without geographical features and those induced using SBR. We conjecture that this may be attributable to the overly simple representation of geographic adjacency used in the sub-representation method of RR-SSA (bin). This is a reasonable conclusion as we can see that using a ``fuller'' representation (i.e.,~RR-SSA (full)) of adjacency produces uniformly improved results.

In examining the different predicted survival curves we have a few observations, summarized as follows. First, we observe that predictive performance increases are mostly realized after the five-year mark. This is, on one hand, intuitive because predicting survival at times closer to the diagnosis is easier than predicting survival at later times. On the other hand, noticeable deviation of the predicted curves uniformly occurs across all models at or around this five-year mark. Therefore, model improvement wrought by using richer geographical representations is realized, by-in-large, at times beyond the five-year mark. Explanation as to \textit{why} such a deviation is present in all models requires further investigation beyond the scope of this work. 

In summary, we find that
\begin{enumerate}
    \item On average, colorectal cancer survival curves can be
reasonably predicted for patients in the state of Iowa.
    \item Geographic features do improve the quality of predicted
colorectal cancer survival curves for patients in the state
of Iowa by 53.5\% (on average) (comparing models induced without geographic features with RR-SSA (full) $k=20$).
    \item On average, RR-SA feature representations improve predictive performance by 15\% over simple representations (SBR) and RR-SSA improve predictive performance by 47.2\% over SBR.
    \item On average, RR-SSA feature representations improve predictive performance by 38.2\% over RR-SA representations (comparing RR-SSA (full) $k=20$ to RR-SA $k=40$).
    \item On average, RR-SSA (full) feature representations improve predictive performance by 32.1\% (comparing RR-SSA (full) $k=20$ to RR-SSA (bin) $k=10$).
\end{enumerate}

\subsubsection{Visualizing Geographic Cluster Assignment}

Next, we briefly discuss the results of visualizing cluster assignment for $k=10,20,30,40$ for RR-SA, RR-SSA (bin), and RR-SSA (full). These results can be observed in Figure \ref{fig:sc_map}, where each unique color represents a single cluster.

For RR-SA (row 1), we first note that as $k$ increases, the elicited geographic regions become more precise, yet maintain geographic continuity. However, we secondly observe that some ZCTAs are not adjacent to any other ZCTA having the same cluster assignment. This disjointedness stems from the use of an adjacency representation of the affinity matrix on which spectral clustering is performed and is not unexpected. As $k$ increases it appears that the number of disjointed ZCTAs also increases. However, we see that the number of continuous regions also increases. In other words, while disjointedness seems to increase with $k$, the desired result of more localized continuous geographical regions is still achieved. Interestingly, when $k=40$, larger Iowa cities such as Des Moines (central Iowa) and Iowa City (central-eastern Iowa) begin to emerge.

In examining row 2 of Figure \ref{fig:sc_map}, representing the RR-SSA (bin) results we see entirely different geographic clusterings than that of RR-SA.  First, we find that for smaller values of $k$ ($k=10,20$, cluster membership is very skewed, with a single cluster dominating the majority of the state, and the remaining cluster assignments being composed of single ZCTAs. These single-ZCTA clusters are found in Des Moines area, the largest urban area of Iowa. As $k$ is increased (i.e., $k=30,40$), rural areas begin to decompose into cluster subsets -- i.e., as the representation is allowed to become more specific (by increasing $k$), rural areas begin to become distinguished between.  Urban areas, such as Des Moines and Iowa City, are also ascribed membership to clusters composed of fewer geographic entities.

Looking at row 3 of Figure \ref{fig:sc_map}, which constitutes the cluster results obtained from RR-SSA (full) models, we observe different clustering results from that of the previous two models. First, we can see that clusters often form ``ring-like'' patterns (this is particularly observable for $k=10$), which is a particularly interesting artifact of this representation. Secondly, juxtaposing these results, with that of the previous two rows (i.e., RR-SA and RR-SSA (bin)), we observe that this representation is somewhat of a ``compromise'' between RR-SA and RR-SSA (full) in the sense that RR-SA produces mostly geographically contiguous clusterings and RR-SSA (bin) produces more geographically disparate clusterings. This is not unexpected, as the sub-representation method of RR-SSA (full) employs the full adjacency representation used in RR-SA, which is not found in RR-SSA (bin). Interestingly, RR-SSA (full) has also discovered urban areas such as Des Moines and Iowa City, but does so at smaller values of $k$ than RR-SSA (bin) (e.g., RR-SSA (bin) $k=10$ is only able to discern areas around Des Moines, whereas RR-SSA (full) is able to discern Iowa City, Des Moines, Waterloo/Cedar Falls, Mason City, etc.). Finally, as $k$ is increased we observe that the representation is becomes more specific in terms of both urban and rural areas up to $k=30$.  When $k=40$ we observe that the clusterings are more disparate, where there appear to be approximately three different rural areas distinguished between (yellow, green, and white), and where urban ZCTAs are assigned to their own unique cluster. This may suggest that urban areas are much more heterogeneous than are rural areas.

\section{Related Work}

The topics related to and discussed throughout this work can best be categorized as \textit{disease and survival curve prediction} and \textit{geographic-based predictions and representation}.

There are many past works involving the prediction of diseases. These can be viewed as classification-based \cite{khosravi2015five,belciug2010two,ojha2017study,sandhuartificial,gupta2011data,belciug2013hybrid,puddu2012artificial} and survival-based \cite{cox1992regression,sharmasurvey,chi2007application, gupta2011data,katzman2016deep,samundeeswari2016artificial}. The focus of this work was on survival curve predictions. Such works can be examined by method, which include Cox proportional hazards model (CPH) \cite{cox1992regression}, which has been historically used to make such predictions, decision trees \cite{sharmasurvey}, and neural network-based models \cite{chi2007application, gupta2011data,katzman2016deep,samundeeswari2016artificial}, which are a more recent development. However, as Laurentiis and Ravdin \cite{de1994technique} point out, CPH has several caveats as compared to neural network-based approaches, including the naivety of the proportional hazards assumption and inability to capture nonlinear feature interactions. Furthermore, decision trees are constructed using greedy methodology and do not have the architectural benefits of neural networks. Hence, this work employed neural networks.

There are also many works focusing on \textit{geographic-based prediction and representation}. These works focus on incorporating geographical features into the predictive process. One method of representing geography is by fine grain lattice (i.e.,~grid) \cite{khezerlou2017traffic,lash2017large,yuan2017predicting}. Such methods are akin to our SBR representation and suffer from the same shortcomings. Spatially adaptive filters \cite{tiwari2005using}, which can tie a single feature to geography when creating $\mathcal{M}$, which may be beneficial when the selected feature is particularly indicative of survival. This method would, however, still produce a binary feature representation, having the accompanying shortcomings discussed when disclosing SBR. Spectral clustering has been used to cluster both social networks \cite{white2005spectral} and for representing geo-spatial features \cite{frias2014spectral,van2013community}, as in this work, and produces a rich (i.e.,~non-sparse) vector of features.

\section{Conclusions and Future Work}

In this work we explored the use of four different geographical feature representations -- a simple binary representation (SBR) and a rich representation based on spectral analysis (which we term spectral analysis and methodologically refer to as RR-SA), and two representations based on similarity-based spectral analysis (RR-SSA) -- to predict colorectal cancer survival curves for patients in the state of Iowa. We show that (a) survival curves can be reasonably estimated, although predictive performance deviates near the five-year survival mark, (b) the use of geographical features generally lead to better predictions, (c) RR-SA trained models outperform those trained using SBR, (d) RR-SSA induced models, generally, outperform RR-SA models, and (e) RR-SSA (full) representations outperform RR-SSA (bin) representations. Future work will involve exploration of different geographical representations, particularly those learned in conjunction with $\mathtt{g}^*$. Additionally, continued exploration of domains and scenarios in which SBR, RR-SA, and RR-SSA geographic representations provide benefit should be undertaken.

\section{Acknowledgements}

The authors would like to thank the Iowa Cancer Registry, State Health Registry of Iowa, and the Iowa Department of Public Health for the data. The authors would also like to thank Gary Hulett and Jason Brubaker for their help in dataset construction and Prakash Nadkarni for his help with both data acquisition and the IRB process.

%

\bibliographystyle{agsm}
\bibliography{CancerReference}

\section*{Appendix}
\setcounter{table}{0}
\renewcommand\thetable{A.\arabic{table}}

\begin{table}[!htp]
    \centering
    \begin{tabular}{ll}
    \toprule
    Model & Avg Optimal Architecture \\\midrule
    No Geo & 1.5:[83,30] \\
    SBR & 1.9:[260,122] \\
    RR-SA, $k=10$ & 1.5:[82,36] \\
    RR-SA, $k=20$ & 1.5:[102,44] \\
    RR-SA, $k=30$ & 1.6:[87,45] \\
    RR-SA, $k=40$ & 1.5:[80,44] \\
    RR-SSA (bin), $k=10$ & 1.6:[82,50] \\
    RR-SSA (bin), $k=20$ & 1.7:[87,50] \\
    RR-SSA (bin), $k=30$ & 1.6:[70,33.33] \\
    RR-SSA (bin), $k=40$ & 1.5:[75,42] \\
    RR-SSA (full), $k=10$ & 1.6:[66,45] \\
    RR-SSA (full), $k=20$ & 1.7:[91,50] \\
    RR-SSA (full), $k=30$ & 1.7:[73,41.43] \\
    RR-SSA (full), $k=40$ & 1.9:[78,42.22] \\\bottomrule
    \end{tabular}
    \caption{Average optimal architecture by model over the 10 folds (e.g., No geo had 1.5 hidden layers, on average, where the first layer had 83 nodes , on average, and the second layer had 30 nodes, on average).\label{tab:params}} 
\end{table}

In Table \ref{tab:params} we can see that, on average, the optimal architecture is relatively comparable among all models with the exception of SBR (and to a degree RR-SA, $k=20$). First, this suggests that the use of RR-SS and RR-SSA features do not affect the architectural complexity of the model. However, SBR seems to significantly increase such complexity. This is somewhat expected, as SBR is represented as a large, sparse vector, which can be contrasted with the comparatively small vector of RR-SA and RR-SSA.

\end{document}